\pdfoutput=1
\documentclass[letterpaper, 10 pt, conference]{ieeeconf}  

\IEEEoverridecommandlockouts                              

\usepackage{graphicx}
\usepackage{cuted}
\usepackage{silence}
\usepackage{caption}
\usepackage{booktabs}   
\usepackage{multirow}
\usepackage{makecell}
\usepackage{amsmath,amssymb}
\usepackage{xcolor}     
\ifdefined\mainonly\usepackage{xr-hyper}\fi 
\usepackage{hyperref}
\usepackage{array}      
\usepackage{placeins}   

\providecommand{\citet}[1]{\cite{#1}}  
\newif\ifshowchanges \showchangestrue
\newif\iffullversion \fullversiontrue
\ifdefined\mainonly\fullversionfalse\fi

\ifdefined\mainonly
  \makeatletter\@addtofilelist{root.aux}\makeatother
  \IfFileExists{root.aux}{}{\typeout{No file root.aux.}}
\fi

\title{\LARGE \bf
Dexterous In-Hand Object Pose Reaching with Learned Safety Filters
}

\author{Ian Chuang$^{1}$, Hiya Shah$^{1}$, Vamsi Chilamkur$^{1}$, Iman Soltani$^{2}$ and Claire Tomlin$^{1}$
\thanks{$^{1}$University of California, Berkeley. $^{2}$University of California, Davis.}
}

\begin{document}

\maketitle
\thispagestyle{empty}
\pagestyle{empty}

\begin{strip}
    \vspace{-15mm}
    \centering
    \captionsetup{type=figure}
    \includegraphics[width=\textwidth]{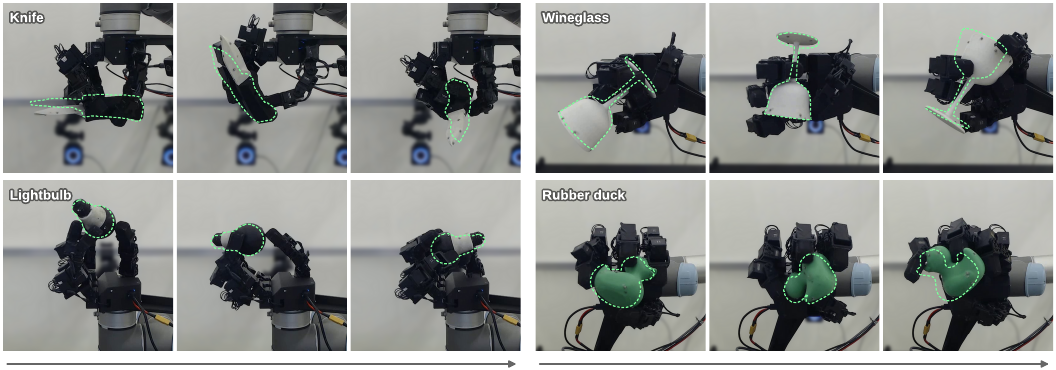}
    
    \vspace{2mm}
    \caption{\textbf{Our robot hand reaches diverse in-hand object poses at any wrist orientation.} For each object (one policy per object), we show three consecutive goals; dashed green outlines indicate target poses, and arrows indicate time.}
    \label{fig:teaser}
\end{strip}


\begin{abstract}
Dexterous manipulation with multi-fingered hands is the next frontier of robotics, yet learned hand skills remain limited to simple grasps or basic repeated motions, such as rotating an object. As robots attempt more demanding in-hand maneuvers, safety, such as not dropping the object, becomes critical, yet it is rarely measured. We study \emph{in-hand object pose reaching}, a task that demands diverse and difficult grasps: moving an object to almost any position and orientation within reach of the fingers, at any wrist orientation, through long sequences of consecutive goals. We present a recipe for training such policies for a LEAP hand in simulation without human data, built on a range of insights, from diverse resets and goals via generated grasp datasets to an adaptive friction curriculum. With careful system identification, the policies transfer zero-shot to the real hand, using either motion capture or a single wrist-mounted depth camera, and prove remarkably capable and robust. On top of the task policy, we add a learned residual safety filter with adjustable conservativeness, letting us tune the balance between performance and safety and greatly reduce object drops. In 13.3 hours of real-robot evaluation over 4{,}800 unseen goals across 4 objects and diverse wrist orientations, the state-based policy reaches 95.2\,\% of goals within 2\,cm and 0.4\,rad, and the filter cuts its drops by 75\,\% while still reaching 91.5\,\% of goals. The policies can also spin a lightbulb held in an arbitrary pose, reach out-of-distribution goals, withstand fast arm motion and perturbations, and, needing only a wrist camera, serve as portable skills such as screwing in a lightbulb. We release the full recipe as a platform for studying performance, reliability, and safety in high-dimensional dexterous systems. Videos and more are available at \url{https://ian-chuang.github.io/dexsafe/}.
\end{abstract}

\section{INTRODUCTION}

Multi-fingered robot hands promise human-like dexterity, but these high-dimensional, contact-rich systems are very difficult to control and leave little margin for error. Recent advances in reinforcement learning (RL) and imitation learning have produced increasingly capable hand controllers, for in-hand reorientation, grasping and everyday manipulation tasks \cite{liu2026dexndm, lum2024dextrahg, ding2025bunny, xu2025dexumi}. Yet the current landscape of in-hand skills remains narrow. Most systems operate at a single wrist orientation, often palm up, where the palm supports the object from below, and perform simple, repetitive motions such as object rotation \cite{handa2023dextreme, qi2022hora, qi2023general}. Many of these works also report performance only over short, limited trials and treat reliability and safety as an afterthought, handled implicitly, e.g., through reward shaping, or not at all. Human in-hand dexterity is far broader: a person can move an object into a difficult grasp at almost any position and orientation in the hand, through diverse finger and wrist motions, and rarely drop it. Achieving such grasps reliably over long sequences of goals remains an open challenge. As robots move outside the lab, where objects can be fragile or hazardous, reliability (how long the system runs without failing), safety (how often it violates a constraint, such as dropping the object) and robustness (to different wrist orientations, scenes and external disturbances) matter as much as success rate. How far a dexterous system can be pushed while staying safe therefore becomes a central question.

To study this question, we consider \emph{in-hand object pose reaching}: moving an object to any position and orientation relative to the hand, at any wrist orientation, over long sequences of consecutive goals (Fig.~\ref{fig:teaser}). Beyond rotation or reorientation \cite{liu2026dexndm, handa2023dextreme, chen2023visual}, the task demands precise translation, so the hand must often hold the object at the tip of its fingers, in hard-to-reach positions and without the palm to support it. Its difficulty makes the task both a frontier for sim-to-real dexterity and a natural testbed for pushing performance and safety together.

We share a complete recipe for training a pose-reaching policy in simulation and transferring it zero-shot to a 16-DoF LEAP hand \cite{shaw2023leap}: goals and initial states drawn from large generated grasp datasets, an adaptive friction curriculum, a simple reward of a few dense goal-reaching terms and regularizers, careful system identification of the actuators \cite{duclusaud2025extended}, and teacher--student distillation into policies that track the object with either motion capture (the \emph{state policy}) or only a wrist-mounted stereo depth camera (the \emph{depth policy}). To improve the policy's safety, we train a residual safety filter, adapted from ResSafe \cite{qu2026ressafe}, that sits on top of the frozen policy and corrects its actions to keep the hand from dropping the object. It trains in a few hours, and a single dial, the correction weight $w$, makes it easy to tune how conservative it is at deployment.

We evaluate the system extensively on a real robot hand: 13.3 hours of robot run time over 4{,}800 unseen, randomly drawn goals, four objects, some of highly nonconvex shape (a knife, a wineglass, a lightbulb and a rubber duck), and six wrist orientations, with and without the safety filter. The state policy reaches 95.2\,\% of goals within 2\,cm and 0.4\,rad, and the filter cuts its drops by 75\,\% (from 16 to 4) while still reaching 91.5\% of goals. Using only the wrist camera, the depth policy still reaches 72\,\% of goals (95\,\% and 88\,\% on the lightbulb and rubber duck), and the filter nearly halves its drops (from 88 to 46). In further experiments, the system reaches out-of-distribution goals, spins a lightbulb held in difficult poses, withstands fast arm motion and physical perturbations, and, since the depth policy needs no external cameras or motion capture, it can easily be deployed with a robot arm as a skill for multi-step tasks, such as screwing in a lightbulb until it lights up, or picking up and reorienting objects.

In summary, our contributions are:
\begin{itemize}
    \item A sim-to-real recipe for robust in-hand object pose reaching with a LEAP hand at any wrist orientation, trained without human data, including a novel adaptive friction curriculum. We ablate the main design choices and will release the code and trained checkpoints.
    \item An adaptation of the learned residual safety filter ResSafe \cite{qu2026ressafe} to dexterous manipulation, as a simple add-on for learning safety in high-dimensional systems that is quick to train and easy to tune, balancing safety and performance.
    \item An extensive real-robot evaluation of performance and safety: 13.3 hours of robot run time and 4{,}800 unseen goals, in which our state policy achieves 95.2\,\% of goals and the safety filter cuts its drops by 75\,\%, and further experiments that stress-test our system with difficult goals and external disturbances and showcase its use as a skill in dexterous arm-and-hand tasks.
\end{itemize}

\section{Related Work}

\textbf{Sim-to-real dexterous manipulation.} Large-scale RL in simulation has produced capable hand controllers \cite{openai2019solving, lum2024dextrahg, kedia2026simtoolreal}, but in-hand work mostly studies rotation around an axis \cite{qi2022hora, qi2023general, liu2026dexndm} or reaching a goal orientation \cite{openai2020learning, handa2023dextreme, chen2023visual, pitz2023dextrous, pitz2024shape}, which do not specify a target position for the object. Few works reach full pose goals on a real hand. TeleDexter \cite{li2026towards} reaches 6D object goals, but it trains on human demonstrations and needs a human reference or operator to supply part of its goal, the fingertip positions; its tasks are shown mostly near a single, palm-down wrist orientation. SimToolReal \cite{kedia2026simtoolreal} moves the arm and the hand together to reach world-frame 6D goals for tool-like objects, relying heavily on the arm to track the goal's position and some of its orientation. POISE \cite{lin2026learning} trains in-hand 6D pose reaching, but its real-robot evaluation covers a few short goal sequences (the longest reported run is 7 consecutive goals) at three wrist orientations, none of them palm down, and its policy acts on an object pose from a pose estimator rather than on visual input. In contrast, our policies use no human data, reach 6D goals at any wrist orientation, and can act end to end from a single wrist depth camera instead of a pose estimator. We further add a learned safety filter, test robustness to disturbances, and evaluate extensively on the real hand: 4{,}800 goals in runs of 50 consecutive goals over 13.3 hours.

\textbf{Safety filters for high-dimensional systems and dexterous manipulation.} Hamilton--Jacobi (HJ) reachability and control barrier functions (CBFs) give safety filters with guarantees \cite{bansal2017hamilton, ames2019control}, but a valid CBF is hard to construct for complex systems \cite{ames2019control}, while classical grid-based HJ scales exponentially with the state dimension and quickly becomes intractable \cite{bansal2017hamilton}. CBF filters exist for grasping and in-hand manipulation, but they rely on simplified models \cite{shawcortez2021control, wu2026safe}. Learned filters avoid an explicit model by learning from data, at the cost of guarantees: a safety critic or HJ value learned with RL \cite{thananjeyan2021recovery, fisac2019bridging} either switches to a fallback policy when the critic deems the state unsafe \cite{nakamura2025generalizing} or steers the nominal action to the closest action that satisfies a CBF-like condition, searching over actions with the critic at run time \cite{oh2025safety, nakamura2026how}. ResSafe \cite{qu2026ressafe} instead learns a residual correction on top of a fixed nominal policy, for high-dimensional humanoid control. We adapt it to in-hand manipulation, where it cuts the state policy's drops by 75\,\%, make its conservativeness a dial that can be tuned at deployment, add a teacher--student distillation stage so that the filter can act on depth observations, and compare it with learned HJ filters. We are not aware of prior work that evaluates a learned safety filter on a real multi-fingered hand, let alone over thousands of goals.

\section{Problem Formulation}
\label{sec:problem}

We define in-hand object pose reaching as the problem of bringing an object to a desired position and orientation relative to the hand. We attach a fixed frame to the object and a fixed frame to the hand (the hand root, below the center of the palm), and express goals in the hand root frame. A goal is either a full $SE(3)$ pose $(p^{*}, R^{*})$ or, for objects that are rotationally symmetric about their $z$-axis (in our case the lightbulb and the wineglass), a goal position $p^{*}$ and a target axis direction $z^{*}$. The axis form leaves the rotation about the object's own axis free, so it can instead be commanded as a spin; this suits objects that look the same at every angle about that axis, whose full pose cannot be recovered from vision alone. With the object at $(p, R)$ in the hand frame, we measure the position and orientation errors as
\begin{equation}
e_p = \lVert p - p^{*} \rVert_2, \quad
e_R = \begin{cases} \lVert \mathrm{Log}(R\,R^{*\top}) \rVert_2 & \text{pose goal},\\
\arccos\!\left(z^{\top} z^{*}\right) & \text{axis goal},\end{cases}
\label{eq:pose_error}
\end{equation}
where $z$ is the object's $z$-axis. For a pose goal, $e_R$ is the geodesic angle on $SO(3)$; for an axis goal, it is the tilt between the object's axis and the target axis. A goal is reached if $e_p$ and $e_R$ are both within a specified tolerance. We perform consecutive goal reaching: goals are commanded one after another, without resetting the hand.

For safety, we consider the constraint of not dropping the object, with the safety margin
\begin{equation}
\ell(x) = d_{\max} - \lVert p_{\mathrm{obj}} - p_{\mathrm{hand}} \rVert_2,
\label{eq:margin}
\end{equation}
where $p_{\mathrm{obj}} - p_{\mathrm{hand}}$ is the offset from the hand root to the object frame. The safe set is $\{x : \ell(x) \geq 0\}$: when the object is more than $d_{\max}$ from the hand root, $\ell(x) < 0$ and we consider the object dropped. We use $d_{\max} = 0.3$\,m for the safety filters and the simulation benchmarks. This is far beyond any goal (all goals lie within about 0.2\,m of the hand root), so leaving the safe set means the object has fallen. The objective is to reach diverse and challenging goals while keeping the system safe, i.e., without dropping the object.

\section{Robot System and Simulation}
\label{sec:system}

\subsection{Hardware}
\label{sec:hardware}

\begin{figure}[t]
    \centering
    \includegraphics[width=\columnwidth]{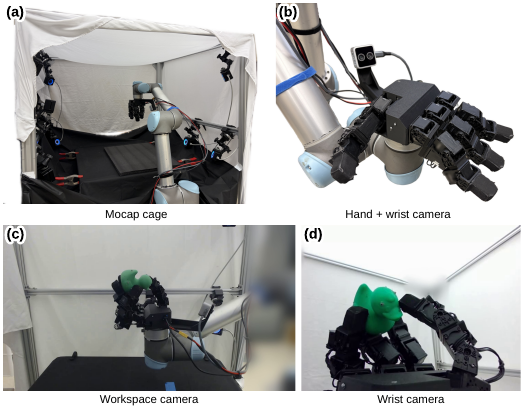}
    \caption{\textbf{Hardware setup.} (a) The UR5 arm with the LEAP hand inside the motion-capture cage. (b) Close-up of the hand and the wrist-mounted stereo camera. (c) Workspace camera view. (d) Wrist camera view.}
    \label{fig:hardware}
\end{figure}

Our setup is shown in Fig.~\ref{fig:hardware}. We use a LEAP hand \cite{shaw2023leap} with a redesigned palm that stabilizes the base joint of the thumb. Its 16 joints are Dynamixel XC330 servos, each running its internal position controller. The fingertips are hard 3D-printed plastic covered with a single layer of high-friction 3M grip tape, and the rest of the hand is mostly low-friction plastic. Visual Dexterity \cite{chen2023visual} found that soft elastomer fingertips make in-hand manipulation more forgiving, but our recipe performs very well without them. The hand is mounted on a UR5 arm, which sets and measures the orientation of the hand, and thus the direction of gravity in the hand frame. The pose-reaching policy controls only the hand, not the arm: it runs at 20\,Hz and sends joint position targets to the servos. Fig.~\ref{fig:objects} shows the four objects: a real lightbulb, and three objects 3D printed in standard, low-friction PLA plastic.

We use a motion-capture system (OptiTrack, 8 cameras) to obtain the object's 6D pose, which is used as input for the state policy. It tracks copies of the objects carrying 10--12 motion-capture markers each, as well as markers on the hand, and gives the object pose in the hand frame. We use a stereo camera on the wrist (Orbbec Gemini 305), which looks at the palm and fingers, for the depth policy. We compute depth from its stereo images with Fast FoundationStereo \cite{wen2026fast} at 30\,Hz, which is much cleaner than the camera's own depth stream, and downsample it to 160$\times$120. The depth policy does not need motion capture to run. In our main experiments, we still run it inside the motion-capture cage on the objects with markers, but only to measure ground truth for scoring goals and drops. Outside the cage, only the depth policy can run, and we use versions of the objects without markers.

For the lightbulb, inside the cage we paint parts of it black, since its reflective surface causes false marker detections. Outside the cage, we instead cover it with tape, since stereo depth on its glossy surface is poor and, unlike paint, the tape still lets the bulb visibly light up when it is screwed in. Neither the paint nor the tape is grippy, so the bulb stays about as slippery as the plastic objects and is not made easier to manipulate. Appendix~\ref{app:hardware} gives the full description of our hardware setup.

\subsection{System Identification}
\label{sec:sysid}
\textbf{Actuator model.} Sim-to-real works commonly simulate each position-controlled joint as a simple PD actuator, which turns the position error directly into a torque, $\tau = k_p(q^{*} - q) - k_d\dot{q}$. Instead, we model the servo itself with BAM \cite{duclusaud2025extended, duclusaud2024bam}, a voltage-driven DC-motor model with an extended friction model: we record our own data on a single-servo pendulum test bench and fit the model's parameters to it, and use these parameters for all 16 actuators of the hand in simulation (Fig.~\ref{fig:sysid}).

\textbf{Latency calibration.} We play the same joint trajectory on the real hand and in simulation, and the time shifts that best align the two give the delays of the joint readings, position commands, and wrist camera. With the identified servo model and delays, the simulated joint positions and depth images closely follow the real ones as shown in Fig.~\ref{fig:sysid_check}. Appendix~\ref{app:sysid} gives the details of actuator and latency system identification.

\subsection{Simulation Environment}
\label{sec:sim_env}
\begin{figure}[t]
    \centering
    \includegraphics[width=\columnwidth]{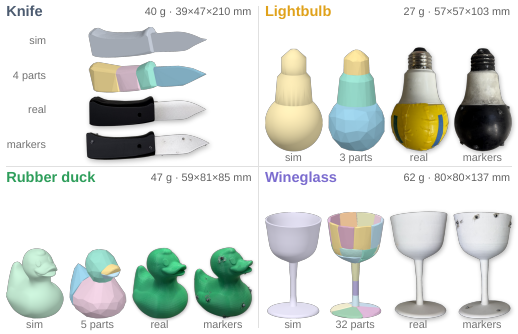}
    \caption{\textbf{Objects used in our experiments.} For each object we show its simulation mesh, its collision meshes from convex decomposition, the real object, and the real object with motion-capture markers.}
    \label{fig:objects}
\end{figure}

We build the simulation in mjlab \cite{zakka2026mjlab}, which runs MuJoCo Warp \cite{googledeepmind2025mujocowarp}, a GPU-accelerated version of MuJoCo \cite{todorov2012mujoco}. The arm is not simulated: the hand is fixed in free space, at a random wrist orientation that changes every episode. We hand-tune our own LEAP hand model, adapted from \cite{ding2025bunny, zakka2022menagerie, shaw2023leap}, adjusting the masses, inertias and collision geometry. Its joints are driven by the identified servo model and delays, and the wrist depth camera is rendered with the calibrated intrinsics and extrinsics of the real camera. Physics runs at 120\,Hz and the policy at 20\,Hz. For collisions, we decompose each object (Table~\ref{tab:objects}, Fig.~\ref{fig:objects}) into convex parts with CoACD \cite{wei2022approximate}. Appendix~\ref{app:sim} gives the full description and the simulator settings. Below we describe the action space, the observations and the domain randomization.

\subsubsection{Action Space}
\label{sec:action}

The action space is inspired by \cite{qi2022hora, li2026towards}. The policy outputs $a_t \in \mathbb{R}^{16}$, and each control step computes
\begin{align}
s_t &= \alpha\,\mathrm{clip}(a_t, -1, 1) + (1-\alpha)\,s_{t-1}, \nonumber\\
\Delta_t &= \kappa\,\mathrm{sign}(s_t)\max(\lvert s_t \rvert - \tau, 0), \nonumber\\
\tilde{q}_t &= \mathrm{clip}(q^{\mathrm{cmd}}_{t-1} + \Delta_t,\ q_t - 0.3,\ q_t + 0.3), \nonumber\\
q^{\mathrm{cmd}}_t &= \mathrm{clip}(\tilde{q}_t,\ q_{\min},\ q_{\max}),
\label{eq:action}
\end{align}
with $\alpha = 0.7$, $\tau = 0.1$ and $\kappa = 0.1$\,rad, where $q_t$ is the measured joint position and $q^{\mathrm{cmd}}$ the position target sent to the servos, initialized to $q_t$ at the start of each episode. Line by line: the action is clipped to $[-1, 1]$ and smoothed by an exponential moving average $s_t$. A deadzone, as in \cite{li2026towards}, turns $s_t$ into the target increment $\Delta_t$: components with $\lvert s_t \rvert \leq \tau$ give no increment, so that small outputs keep the hand still instead of drifting. The increment is added to the previous target, which is kept within 0.3\,rad of the measured position so that $q^{\mathrm{cmd}}$ cannot drift far from the actual joint. Finally, the target is clipped to the joint limits. Each joint therefore moves its target by at most 0.09\,rad per step (1.8\,rad/s).
We found the clipping to $[-1, 1]$ together with the small scale of 0.1\,rad to be essential to keep the hand from moving too fast, since it limits the maximum speed of the motors. Appendix~\ref{app:action} further discusses this choice of action space.

\subsubsection{Observations}
\label{sec:obs}
The policies draw on the observations of Table~\ref{tab:observations}, grouped by the symbols of Fig.~\ref{fig:method}. The proprioception $o^{\mathrm{prop}}$ is what is also available on the real robot: the gravity direction in the hand frame, a 0.5\,s history of joint positions, the residual between the commanded target $q^{\mathrm{cmd}}$ and the measured position $q$, and the last action. $x$ is the object pose (from motion capture on the robot), $g$ the goal, and $I$ the 160$\times$120 wrist depth image. The privileged observations $o^{\mathrm{priv}}$ add noise-free positions and velocities of the joints, fingertips and object, and fingertip contacts. A few further terms, such as the fingertip friction coefficients and the success bookkeeping, are given only to the critic (value function) in RL training (Sec.~\ref{sec:stages}), as in asymmetric actor--critic training \cite{pinto2018asymmetric}. These terms help the critic predict returns, but we keep them from the actor, which could exploit them and learn undesired behavior.

To narrow the visual gap, we augment the simulated depth following DextrAH-G \cite{lum2024dextrahg}: pixel jitter, disparity noise, surface-normal-dependent noise, pixel dropout, random-value pixels and thin stick artifacts. We also randomize the camera pose and field of view, and when training the depth policies we add random floating distractor objects around the hand, so that the depth policy becomes robust to visual distractors (Table~\ref{tab:domain_randomization}, Fig.~\ref{fig:depth_sim}).

\subsubsection{Domain Randomization}
\label{sec:dr}
To transfer to the real robot, we randomize the simulation broadly: friction, actuator parameters, masses and inertias, and collision geometry, as well as random pushes on the object, delayed and dropped commands, and noise and latency on the observations the robot provides (Appendix~\ref{app:dr}).

\section{Learning Pose Reaching and Safety}
\label{sec:overview}

\begin{figure*}[t]
    \centering
    \includegraphics[width=\textwidth]{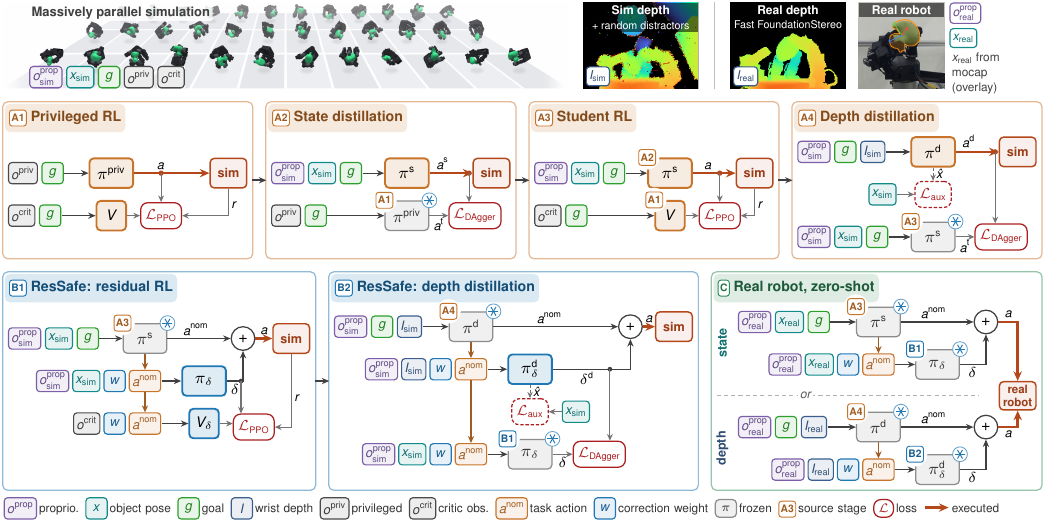}
    \caption{\textbf{Training pipeline.} A1--A4: The task policy: A1 PPO with privileged observations and an asymmetric critic, A2 DAgger into a deployable state student, A3 PPO fine-tuning of that student, and A4 DAgger into a depth student. B1--B2: The safety filter: B1 learns a residual safety filter on top of the frozen A3 policy; B2 distills it into a depth student filter on top of A4. C: This gives a task policy and a safety filter that use either motion capture or the wrist camera, and both transfer zero-shot to the real robot.}
    \label{fig:method}
\end{figure*}

Our training pipeline is outlined in Fig.~\ref{fig:method}. It has two parts. The pose-reaching policy, which we also call the task policy, is trained in four stages (A1--A4), and the safety filter in two stages (B1--B2) on top of the frozen task policies. Every stage trains one network per object.
For each object we obtain two task policies: the \emph{state policy} ($\pi^{\mathrm{s}}$, A3), which observes the object pose from motion capture, and the \emph{depth policy} ($\pi^{\mathrm{d}}$, A4), which sees only the wrist depth camera; and a safety filter for each, the state filter (B1) and the depth filter (B2). Sec.~\ref{sec:task_policy} describes the task policies, and Sec.~\ref{sec:ressafe} the safety filter.

\subsection{Pose-Reaching Policy}
\label{sec:task_policy}
We train the pose-reaching policies with RL, using PPO \cite{schulman2017proximal} in thousands of parallel simulated environments. Below we describe the RL environment built on the simulation of Sec.~\ref{sec:sim_env}: how we set up goals and initial states, the reward, the curricula and the training stages.

\subsubsection{Goals and Resets from a Grasp Dataset}
\label{sec:goals}
Rather than sampling object pose goals uniformly over $SE(3)$, where many poses cannot be held by the hand, we generate a grasp dataset for each object and our hand offline, with no human data, so that each goal is the object pose of a grasp and thus more likely reachable. We use LightningGrasp \cite{yin2025lightning}, configured for our modified LEAP hand, to produce 100{,}000 hand-object grasps. We then expand them into 500{,}000 grasps whose object poses densely cover the space above the palm, following the RRT formulation of DexterityGen \cite{yin2025dexteritygen}. Finally, we remove grasps with collisions between the object and hand geometry, which leaves 422k--453k grasps per object (Appendix~\ref{app:grasp}, Fig.~\ref{fig:grasp_dataset}). Each grasp stores the 16 joint angles and the object pose in the hand frame. Goals and initial states are drawn from this dataset.

\textbf{Resets and terminations.} Recent works \cite{yin2025dexteritygen, liao2026beyondmimic, lin2026learning, li2026towards, yin2026emergent} use targeted, diverse resets to guide RL exploration. Inspired by these works, at every reset we start the hand and object in a random grasp drawn from our grasp dataset, with the hand at a uniformly random wrist orientation. An episode times out at 20\,s and terminates early when the object falls more than 0.5\,m from the hand (Appendix~\ref{app:resets}).

\textbf{Goal completion in training.} In training, we use a stricter condition to decide when a goal is complete and the next one is drawn, so that the policy learns to settle the object at the goal rather than pass through it: the object must be within three tolerances, and stay there. The position and orientation errors \eqref{eq:pose_error} must be below $\tau_p$ and $\tau_R$. The object must also move at the goal twist $\omega^{*}$: we low-pass filter the object's angular velocity $\omega$ in the hand frame,
\begin{equation}
\bar{\omega}_t = (1-\alpha_\omega)\,\bar{\omega}_{t-1} + \alpha_\omega\,\omega_t, \qquad e_\omega = \lVert \bar{\omega}_t - \omega^{*} \rVert_2,
\label{eq:twist}
\end{equation}
with $\alpha_\omega = 0.05$, and require the twist error $e_\omega$ to be below $\tau_\omega$. For the lightbulb, the goal twist is a spin about the goal axis, $\omega^{*} = \omega^{*}_z z^{*}$ with $\omega^{*}_z \sim \mathcal{U}(-1.5, 1.5)$\,rad/s (or zero with probability 0.5); for every other object it is zero. This makes the policy hold the object still, or spin it steadily, at the goal. Finally, the goal must be held: a hold counter, reset to zero for each new goal, goes up by one at every step within all three tolerances and down by one (but not below zero) at every step outside them, and the goal is complete when the counter reaches the required hold $h_{\mathrm{req}}$, drawn per goal. If the object briefly slips outside the tolerances, this only delays completion without restarting the count. Table~\ref{tab:objects} lists $\tau_p$, $\tau_R$, $\tau_\omega$ and the range of $h_{\mathrm{req}}$ for each object. A new goal is drawn at every reset and when the current one is complete, so the policy learns to reach goals consecutively.

\subsubsection{Reward}
\label{sec:reward}
We keep the reward simple: dense goal-reaching terms and a few regularizers. We write the goal-reaching terms with potential-based shaping \cite{ng1999policy}: a potential $\Phi$ measures progress toward the goal, such as the negative distance to it, and the reward is its change between steps, $\Delta\Phi_t = \Phi_t - \Phi_{t-1}$. Moving toward the goal is rewarded, moving away is penalized by the same amount, and hovering near the goal without reaching it earns nothing. We prefer this to rewards for closeness to the goal \cite{yin2025dexteritygen, lin2026learning, li2026towards}, which pay at every step near the goal and so can be exploited by hovering there. The task reward is
\begin{align}
r_t = \Delta t\,\Big[&\,600\,\Delta\Phi^{p}_t + 25\,\Delta\Phi^{R}_t + 10\,\Delta\Phi^{\omega}_t + 35\,\Delta\Phi^{h}_t \nonumber\\
&- 50\,\mathbf{1}_{\mathrm{drop}} + 0.01\,(n^{\mathrm{tip}}_t - n^{\mathrm{link}}_t - n^{\mathrm{self}}_t) \nonumber\\
&- 0.01\,\lVert a_t - a_{t-1} \rVert^2 - 0.005\,\lVert a_t \rVert^2 \Big],
\label{eq:reward}
\end{align}
with the potentials
\begin{gather}
\Phi^{p} = -\min(e_p,\ 0.1\,\mathrm{m}), \qquad \Phi^{R} = -e_R, \qquad \Phi^{h} = h / h_{\mathrm{req}}, \nonumber\\
\Phi^{\omega} = \begin{cases} -\min(e_\omega,\ 1.5) & \text{if } e_p < \tau_p \text{ and } e_R < \tau_R,\\ -1.5 & \text{otherwise.}\end{cases}
\end{gather}
The position and orientation potentials $\Phi^{p}$ and $\Phi^{R}$ are the negative errors to the goal, the position capped at 10\,cm. The twist potential $\Phi^{\omega}$ is the negative error to the goal twist, counted only once the pose is within tolerance. The hold potential $\Phi^{h}$ rises from 0 to 1 as the hold counter $h$ fills. The drop penalty $\mathbf{1}_{\mathrm{drop}}$ is applied if the episode terminates with a drop. The contact terms count fingertips touching the object ($n^{\mathrm{tip}}$), other links touching it ($n^{\mathrm{link}}$) and self-contacts ($n^{\mathrm{self}}$), which encourages fingertip grasps and discourages non-fingertip contact and self-collision. The last two terms penalize the action rate and magnitude, for smoother motion. Finally, as in mjlab \cite{zakka2026mjlab}, the sum is scaled by the control period $\Delta t = 0.05$\,s. Appendix~\ref{app:reward} provides more details.

\subsubsection{Friction and Goal Curricula}
\label{sec:curriculum}
Curricula speed up learning dramatically. When training the privileged teacher (A1), we use a friction curriculum and a goal curriculum, both adaptive in the style of the curriculum in the OmniReset code release \cite{yin2026emergent}: each has a progress $p \in [0, 1]$ that steps up when the fraction of completed goals over the last 500 goals is above a threshold and down when it falls below a slightly lower one. The friction curriculum starts with unrealistically high friction ($\mathcal{U}(2.5, 3.0)$ at the fingertips) and moves it linearly to the final range ($\mathcal{U}(0.3, 1.2)$). It changes the difficulty smoothly without changing the structure of the task. The goal curriculum builds on the goal rule of DexterityGen \cite{yin2025dexteritygen}, where the next goal is one of the 10 grasps nearest to the current state among a random subset of the dataset. We extend this rule into a curriculum that gradually moves goals from near to far as the policy improves over training: we shrink this subset from 2048 grasps to 1 as $p$ rises, so goals move from near the current grasp to uniform over the dataset, while the success tolerances $\tau_p$, $\tau_R$ and $\tau_\omega$ tighten and the required hold $h_{\mathrm{req}}$ grows. All later stages run without curricula, at their final settings (low friction and goals uniform over the dataset). We ablate both curricula and also a gravity curriculum, which starts with reduced gravity and ramps it to its full value \cite{li2026towards, chen2022system}: the friction curriculum speeds up learning the most on three of the four objects, while the goal and gravity curricula help little on their own (Sec.~\ref{sec:abl_curric}). Appendix~\ref{app:curric} gives the full description and the schedules.

\subsubsection{Training Stages}
\label{sec:stages}
We follow the common recipe of distilling privileged teachers into deployable students. A1 trains a teacher with PPO \cite{schulman2017proximal, schwarke2025rslrl} on privileged observations ($o^{\mathrm{priv}}, g^{\mathrm{priv}}$) with an asymmetric critic; it is the only stage that uses the friction and goal curricula. We find that A1 learns far better with a privileged actor: an actor without the privileged observations, which would be deployable on the real robot as in \cite{li2026towards, kedia2026simtoolreal}, reaches about half as many goals with the same amount of training (Sec.~\ref{sec:abl_priv}). A2 distills the teacher with online DAgger \cite{ross2011reduction} into a student that uses only observations available on the real robot ($o^{\mathrm{prop}}, x, g$), including the object pose from motion capture. A3 fine-tunes this student with PPO in the same environment, which recovers part of the performance lost in the A2 distillation; its critic is copied from A1. A4 distills A3 into a student that replaces the object pose with the wrist depth image, encoded by a CNN with a spatial softmax \cite{levine2016end}. An auxiliary head and loss predict the object pose from the same features to help the encoder localize the object \cite{lum2024dextrahg}. A4 trains at relaxed success tolerances ($\tau_p = 2$\,cm, $\tau_R = 0.4$\,rad), with visual distractors around the hand and a slightly randomized camera pose and field of view (Sec.~\ref{sec:obs}). The results are the state policy (A3) and the depth policy (A4). On one machine with RTX 5090 GPUs, A1 takes about 15--17\,h on two GPUs (28\,h for the wineglass), and A2, A3 and A4 take 0.6, 2.5--2.8 and 10\,h on one GPU. Appendix~\ref{app:training} gives the full description and the hyperparameters.

\subsection{Residual Safety Filter}
\label{sec:ressafe}

On top of the frozen task policies, we train a safety filter (Fig.~\ref{fig:method}, B1--B2). We follow ResSafe \cite{qu2026ressafe}, which showed that a modular safety filter can be trained on top of a nominal policy for high-dimensional systems that are hard to model: a learned residual keeps the system alive, i.e., safe, while being encouraged to deviate as little as possible from the nominal policy. This separates learning the task from learning safety. Making the task policy itself safer, e.g., with a larger drop penalty, turns its reward into a multi-objective problem whose trade-off is hard to tune and requires retraining the task policy (about 28--30\,h per object, 42\,h for the wineglass). The filter is instead a quick layer on top of the frozen policy (about 4\,h of training), and it is easy to tune and deploy. 

\subsubsection{Formulation}
\label{sec:ressafe_formulation}
Let $a^{\mathrm{nom}}_t = \mathrm{clip}(\pi(\cdot), -1, 1)$ be the action of the frozen task policy. The safety policy $\pi_\delta$ observes $a^{\mathrm{nom}}_t$ and outputs a residual $\delta_t$, and the executed action is
\begin{equation}
a_t = \mathrm{clip}\!\left(a^{\mathrm{nom}}_t + \delta_t,\ -1,\ 1\right),
\label{eq:residual}
\end{equation}
which then passes through the same action law \eqref{eq:action} as the task policy. The clip keeps the executed action within its bounds whatever the residual. It does not change what the robot does, since the action law clips to $[-1, 1]$ anyway. ResSafe trains $\pi_\delta$ to maximize the discounted return $\mathbb{E}\big[\sum_{t=0}^{T-1} \gamma^t r_t\big]$, where $T$ is the step at which the system leaves the safe set, with a per-step reward $r_t = 1 - \lambda\lVert\delta_t\rVert^2$: each step alive earns one unit, minus a penalty on the correction. We keep this objective but change the reward so that it is bounded and never negative,
\begin{equation}
r_t = \frac{1}{1 + w\,\lVert \delta_t \rVert_2^2 / n},
\label{eq:ressafe_reward}
\end{equation}
where $n = 16$ is the action dimension, so that the scale of $w$ does not depend on the robot. With an unbounded negative reward, a policy can prefer to fail as fast as possible to stop accruing negative reward (the original ResSafe counters this with a large termination penalty); with $r_t \in (0, 1]$, every step alive is worth something, and an uncorrected step is worth the most. When the object leaves the safe set ($\ell(x) < 0$, i.e., more than 0.3\,m from the hand), the episode terminates with no further reward and the environment is reset.

Unlike the original ResSafe, which adds environment-specific terms (rewarding a steady humanoid base) and penalizes the action rate and termination, we add no environment-specific reward terms. The reward is exactly \eqref{eq:ressafe_reward}, which keeps the formulation simple and general.

\subsubsection{Run-Time Conservativeness and Training}
\label{sec:ressafe_w}
Unlike the original ResSafe, which trains with a single fixed correction weight, we make $w$ an input. We draw $w \sim \mathcal{U}(0, 2)$ per environment and episode and give it to both actor and critic as an extra observation, so one network covers the whole range and $w$ becomes a dial at deployment: a large $w$ keeps the executed action close to the nominal policy, and $w = 0$ lets the filter intervene freely.

The filter is an asymmetric actor--critic: the actor sees only deployable observations, the proprioception and object pose $(o^{\mathrm{prop}}, x)$, plus $a^{\mathrm{nom}}$ and $w$, and the critic sees the privileged state in place of $(o^{\mathrm{prop}}, x)$; neither sees the goal. Since the task policy is already trained, the filter trains in the task environment at its final curriculum settings, with no curricula of its own. Also unlike the original, we add Gaussian noise of a random per-episode scale to the nominal action during training, so the filter sees more diverse nominal actions.

B1 trains the state filter $\pi_\delta$ with PPO for 4000 iterations on top of the state policy (2.5\,h on one RTX 5090), and B2 distills it with online DAgger into the depth filter $\pi^{\mathrm{d}}_\delta$, which runs on top of the depth policy (1.8\,h). At deployment we use $w = 0.5$, chosen from the simulated sweep over $w$ (Sec.~\ref{sec:sim_results}), unless stated otherwise. We train three B1 seeds per object and deploy the one with the highest goal success at each object's training goal tolerance (Table~\ref{tab:objects}), averaged over $w$ in simulation (Appendix~\ref{app:seeds}). Appendix~\ref{app:ressafe_train} gives further training details.

\subsubsection{Baseline: Hamilton--Jacobi Safety Filters}
\label{sec:hj_baseline}
We also compare with another family of learned safety filters, which combines HJ reachability with RL. These filters learn a safety critic $Q(x, u)$, which predicts how safe the system stays after taking action $u$ in state $x$, and a fallback policy $\pi_{\text{safe}}$, which outputs the action the critic predicts is safest. The critic is trained with the discounted safety Bellman backup of \cite{fisac2019bridging}. They are used at run time in one of two ways. The least-restrictive safety filter (LRSF), as in \cite{nakamura2025generalizing}, executes the task action until the critic predicts a violation and then switches to the fallback action. The human-centered safety filter (HCSF) \cite{oh2025safety, nakamura2026how} instead executes the action closest to the task action that satisfies a discrete-time CBF condition, $Q(x, u) \geq \beta\, Q(x, \pi_{\text{safe}}(x))$, where $\beta \in [0, 1)$ sets how early and smoothly the filter brakes (a larger $\beta$ intervenes earlier). The action is found by a sampling-based line search over actions between the nominal and fallback actions. These filters are trained with off-policy actor--critic methods such as SAC \cite{haarnoja2018soft}, which are not designed for thousands of parallel simulated environments, so we train them with FlashSAC \cite{kim2026flashsac}. Its critic is distributional: it predicts a distribution over the safety value. Instead of the mean of this distribution, we use its conditional value at risk $\mathrm{CVaR}_\alpha$, a pessimistic estimate, so $\alpha$ (and $\beta$ for HCSF) tunes how conservative the filter is, i.e., how easily it intervenes. Appendix~\ref{app:hj} gives the background, the equations and our implementation.

\section{Experiments}
\label{sec:experiments}

\begin{figure}[t]
    \centering
    \includegraphics[width=\columnwidth]{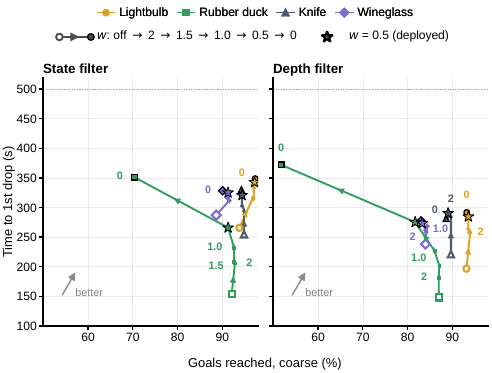}
    \caption{\textbf{Effect of the correction weight $w$ in simulation.} Coarse goal success vs.\ time to the first drop, per object, for the state (left) and depth (right) policies, from no filter (hollow) through $w = 2$ to 0.}
    \label{fig:sim_w_dial}
\end{figure}

In our experiments we want to evaluate performance, reliability and safety extensively, stress-testing the system over many trials and external disturbances.
We evaluate in simulation and on the real robot: (i) how well the task policies reach goals, how often they drop the object, and how performance and safety interact, e.g., whether the filter lets the robot reach more goals before a drop (Secs.~\ref{sec:sim_results} and~\ref{sec:real_main}); (ii) how ResSafe compares with HJ safety filters (Sec.~\ref{sec:hj_results}); (iii) which design choices matter (Sec.~\ref{sec:ablations}); and (iv) further real-robot experiments with difficult goals, disturbances and downstream tasks that use the pose-reaching policy as a skill (Secs.~\ref{sec:twist}--\ref{sec:tasks}).

\subsection{Simulation Results}
\label{sec:protocol}
\textbf{Protocol and metrics.} The simulation benchmark runs 1{,}024 environments, each with one random wrist orientation held for 50 consecutive goals drawn from the training goal distribution, with a 10\,s window per goal. We command no twist for the lightbulb here and test spinning separately (Sec.~\ref{sec:twist}). A goal counts as reached if $e_p$ and $e_R$ are both within the tolerance at any step of its 10\,s window, without the twist and hold conditions of training. We use two tolerances for every object: \emph{coarse} (2\,cm, 0.4\,rad) and \emph{precise} (1\,cm, 0.2\,rad), with the tilt angle as the orientation error for the lightbulb and wineglass. A drop is the object leaving the 0.3\,m safe set; the environment is then reset to a fixed grasp and continues with the next goal. We report the percentage of goals reached and dropped, the mean time to an environment's first drop (500\,s if it never drops), and the RMS correction of the filter. Appendix~\ref{app:eval} gives the full protocol.

\subsubsection{Performance, Safety and the Correction Weight}
\label{sec:sim_results}

For each object, we run the task policy without and with the filter at $w \in \{0, 0.5, 1, 1.5, 2\}$ (Fig.~\ref{fig:sim_w_dial}). For both the state and the depth policy, the filter delays the first drop on every object, and lowering $w$ moves each object along a trade-off curve: $w = 0$ intervenes hardest (it generally drops latest and reaches the fewest goals) while $w = 2$ stays close to the nominal policy. For the knife, lightbulb and wineglass the curves are flatter, i.e., the filter removes many drops at little cost in success, whereas the rubber duck pays the most for safety. Based on these results, we choose $w = 0.5$ for the real robot as a middle ground: it removes most of the drops that $w = 0$ removes, at a much smaller cost in success. Appendix~\ref{app:results} gives the full metrics (Table~\ref{tab:sim_results}).

\subsubsection{Comparison with HJ Safety Filters}
\label{sec:hj_results}
Table~\ref{tab:filter_baselines} compares ResSafe with the HJ filters (Sec.~\ref{sec:hj_baseline}) on the rubber duck with the state policy, showing three HCSF settings from a sweep over $\alpha$ and $\beta$ with drop rates near ResSafe's. ResSafe at $w = 0.5$ halves drops while staying within one point of the unfiltered policy's coarse success. At a similar drop rate, HCSF reaches 9 points fewer goals, and the hard-switching LRSF reaches even fewer coarse goals with more drops. ResSafe is also a single forward pass whose trade-off is set by one input, whereas the HJ filter needs a run-time search over actions and is sensitive to $\alpha$ and $\beta$. Appendix~\ref{app:hj_impl} gives the full sweep and the training details.

\begin{table}[t]
\centering
\scriptsize
\setlength{\tabcolsep}{2.2pt}
\caption{ResSafe vs.\ the HJ filters in simulation (rubber duck, state policy, 1{,}024 environments $\times$ 50 goals): three HCSF settings near ResSafe's drop rate (full sweep in Table~\ref{tab:filter_baselines_sweep}) and LRSF at $\alpha{=}0.1$. Mean $\pm$ standard error of the mean over the 1{,}024 environments. }
\label{tab:filter_baselines}
\resizebox{\columnwidth}{!}{%
\begin{tabular}{@{}ll ccc c c@{}}
\toprule
 & & \multicolumn{2}{c}{\makecell{Goals reached (\%)\,$\uparrow$}} & & \makecell{Time to 1st\\drop or} & \makecell{Avg.\\action\\change\\(\%)} \\
\cmidrule(lr){3-4}
Obs. & Filter & coarse & precise & Drops (\%)\,$\downarrow$ & end (s)\,$\uparrow$ &  \\
\midrule
\multirow{6}{*}{\rotatebox[origin=c]{90}{State}} & None & 92.1\,{\tiny$\pm$0.3} & 88.4\,{\tiny$\pm$0.3} & 9.1\,{\tiny$\pm$0.3} & 154\,{\tiny$\pm$5} & -- \\
 & ResSafe, $w{=}0.5$ & 91.3\,{\tiny$\pm$0.3} & 80.1\,{\tiny$\pm$0.4} & 4.5\,{\tiny$\pm$0.3} & 266\,{\tiny$\pm$6} & 15.3 \\
 & HCSF $\alpha{=}0.3,\beta{=}0.9$ & 87.4\,{\tiny$\pm$0.5} & 80.4\,{\tiny$\pm$0.5} & 6.6\,{\tiny$\pm$0.2} & 185\,{\tiny$\pm$5} & 10.8 \\
 & HCSF $\alpha{=}0.2,\beta{=}0.9$ & 84.7\,{\tiny$\pm$0.4} & 75.3\,{\tiny$\pm$0.5} & 6.0\,{\tiny$\pm$0.2} & 201\,{\tiny$\pm$5} & 14.4 \\
 & HCSF $\alpha{=}0.3,\beta{=}0.95$ & 82.0\,{\tiny$\pm$0.4} & 70.8\,{\tiny$\pm$0.5} & 4.8\,{\tiny$\pm$0.1} & 230\,{\tiny$\pm$6} & 17.2 \\
 & LRSF $\alpha{=}0.1$ & 79.3\,{\tiny$\pm$0.6} & 71.3\,{\tiny$\pm$0.7} & 8.2\,{\tiny$\pm$0.2} & 154\,{\tiny$\pm$5} & 16.5 \\
\bottomrule
\end{tabular}}
\end{table}

\subsubsection{Ablations}
\label{sec:ablations}
We ablate the main design choices in simulation: the curricula, the privileged actor observations of A1, and two design choices of the filter.

\begin{figure}[t]
    \centering
    \includegraphics[width=\columnwidth]{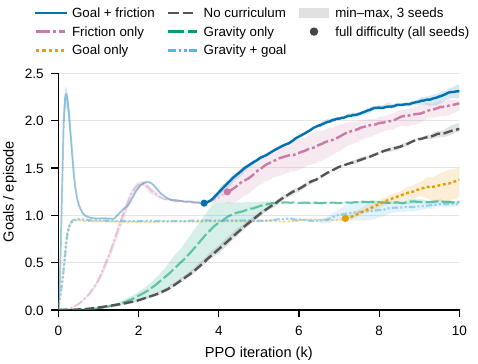}
    \caption{\textbf{Knife curriculum ablation over 3 seeds.} Goals per episode during A1 for each curriculum variant, including a gravity curriculum (0$\to$9.81\,m/s$^2$). Mean with min--max band; dots mark full difficulty. Gravity runs are faded, since most do not reach full gravity within 10k iterations.}
    \label{fig:knife_curric_ablation}
\end{figure}

\label{sec:abl_curric}
\textbf{Curricula.} On the knife, the friction curriculum contributes the most to A1 (Fig.~\ref{fig:knife_curric_ablation}, 3 seeds): goal + friction reaches 2.27 goals per episode, friction only 2.14, no curriculum 1.87 and goal only 1.33. A gravity curriculum \cite{li2026towards, chen2022system} does not help in our setting (1.11--1.14 goals per episode). Appendix~\ref{app:abl} gives additional curriculum ablation results on the other objects where including the friction curriculum generally helps the most.

\label{sec:abl_priv}
\textbf{Privileged actor observations.} Training the A1 actor on privileged observations is essential (Sec.~\ref{sec:stages}): on the knife, an actor that sees only the student's noisy, delayed observations plateaus at 1.12 goals per episode against 2.27 for the privileged actor, over 3 seeds each (Fig.~\ref{fig:training_observation_ablation}).

\label{sec:abl_ressafe}
\textbf{ResSafe design.} On the rubber duck, over 3 seeds, an L1 correction penalty gives a worse trade-off than the squared L2 penalty at every $w$ (Fig.~\ref{fig:ressafe_ablation}): at $w = 0.5$, it reaches only 21.7\,\% of goals at the coarse tolerance, against 80.4\,\% with L2 (means over the three seeds). Removing the noise on the nominal action during training does not significantly change safety or performance in simulation. We keep the noise because it diversifies the nominal actions the filter sees (Appendix~\ref{app:abl}).

\subsection{Real Robot Results}

\subsubsection{Performance and Safety over 4{,}800 Goals}
\label{sec:real_main}

\begin{figure}[t]
    \centering
    \includegraphics[width=\columnwidth]{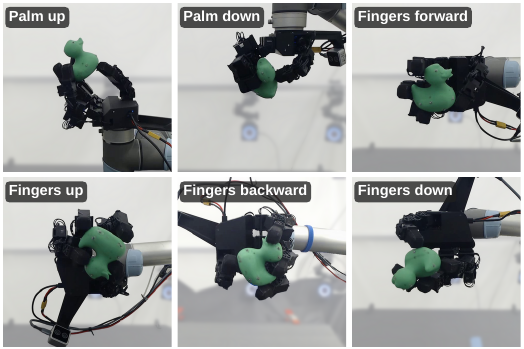}
    \caption{\textbf{The six evaluation wrist orientations}, which point gravity along each axis of the hand.}
    \label{fig:wrist_orientations}
\end{figure}

\begin{table}[t]
\centering
\scriptsize
\setlength{\tabcolsep}{2.2pt}
\caption{Real-robot results. Each row pools 6 runs of 50 consecutive goals (one per wrist orientation), 300 goals. Precise and coarse: goals reached within 1\,cm, 0.2\,rad and 2\,cm, 0.4\,rad. Drops: goals in which the object fell. Time to the first drop (or 500\,s, the end of the run) and coarse goals reached before the first drop: means over the 6 runs. Last column: RMS filter correction (\% of the action limit). Bold: better of each pair. The filter cuts drops at a small cost in success.}
\label{tab:real_main_col}
\resizebox{\columnwidth}{!}{%
\begin{tabular}{@{}lll ccccc c@{}}
\toprule
 & & & \multicolumn{2}{c}{\makecell{Goals reached\\(/300)\,$\uparrow$}} & \makecell{Drops\\(/300)\\$\downarrow$} & \makecell{Time to\\1st drop\\or end\\(s)\,$\uparrow$} & \makecell{Reached\\before 1st\\drop (/50)\,$\uparrow$} & \makecell{Avg.\\action\\change\\(\%)} \\
\cmidrule(lr){4-5}
Obs. & Object & \makecell{Safety\\filter} & precise & coarse & & & &  \\
\midrule
\multirow{8}{*}{\rotatebox[origin=c]{90}{State}} & \multirow{2}{*}{Lightbulb} & $\times$ & \textbf{277} & \textbf{297} & 2 & 427 & 42.3 & -- \\
 &  & $\checkmark$ & 268 & 292 & \textbf{0} & \textbf{500} & \textbf{48.7} & 8.5 \\
\cmidrule(l){2-9}
 & \multirow{2}{*}{Rubber duck} & $\times$ & \textbf{255} & \textbf{291} & 7 & 298 & 29.3 & -- \\
 &  & $\checkmark$ & 194 & 262 & \textbf{2} & \textbf{441} & \textbf{39.2} & 15.4 \\
\cmidrule(l){2-9}
 & \multirow{2}{*}{Knife} & $\times$ & \textbf{234} & \textbf{288} & 4 & 257 & 24.7 & -- \\
 &  & $\checkmark$ & 216 & 281 & \textbf{1} & \textbf{470} & \textbf{44.5} & 9.3 \\
\cmidrule(l){2-9}
 & \multirow{2}{*}{Wineglass} & $\times$ & \textbf{141} & \textbf{266} & 3 & 352 & 30.8 & -- \\
 &  & $\checkmark$ & 128 & 263 & \textbf{1} & \textbf{449} & \textbf{39.2} & 8.9 \\
\midrule
\multirow{8}{*}{\rotatebox[origin=c]{90}{Depth}} & \multirow{2}{*}{Lightbulb} & $\times$ & \textbf{191} & \textbf{284} & 3 & 371 & 35.2 & -- \\
 &  & $\checkmark$ & 183 & 278 & \textbf{0} & \textbf{500} & \textbf{46.3} & 7.4 \\
\cmidrule(l){2-9}
 & \multirow{2}{*}{Rubber duck} & $\times$ & \textbf{152} & \textbf{264} & 12 & 268 & 23.5 & -- \\
 &  & $\checkmark$ & 95 & 233 & \textbf{2} & \textbf{392} & \textbf{30.3} & 13.1 \\
\cmidrule(l){2-9}
 & \multirow{2}{*}{Knife} & $\times$ & \textbf{60} & \textbf{163} & 22 & 163 & 9.0 & -- \\
 &  & $\checkmark$ & 53 & 146 & \textbf{4} & \textbf{406} & \textbf{22.0} & 6.6 \\
\cmidrule(l){2-9}
 & \multirow{2}{*}{Wineglass} & $\times$ & \textbf{56} & \textbf{151} & 51 & 111 & 3.7 & -- \\
 &  & $\checkmark$ & 50 & 143 & \textbf{40} & \textbf{117} & \textbf{5.5} & 6.6 \\
\bottomrule
\end{tabular}}
\end{table}

\begin{figure}[t]
    \centering
    \includegraphics[width=\columnwidth]{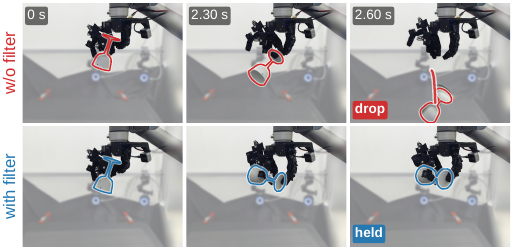}
    \caption{\textbf{ResSafe prevents a drop.} The same wineglass goal (palm down, depth policy) without the filter (top), where the glass slips and falls, and with it (bottom). The red line traces the falling glass.}
    \label{fig:wineglass_filter_save}
\end{figure}

\begin{figure}[t]
    \centering
    \includegraphics[width=\columnwidth]{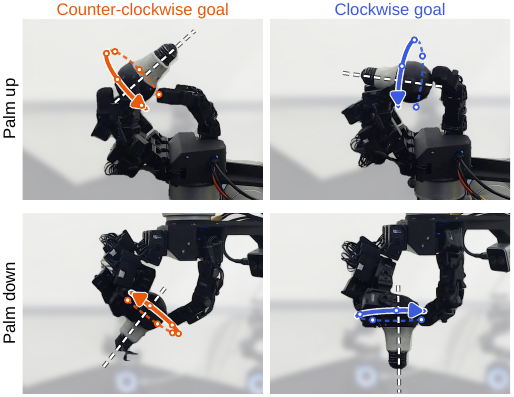}
    \caption{\textbf{Spinning the lightbulb about its own axis} (depth policy). Palm up (top) and down (bottom), counter-clockwise (left) and clockwise (right) goals. }
    \label{fig:lightbulb_twist}
\end{figure}

\begin{figure}[t]
    \centering
    \includegraphics[width=\columnwidth]{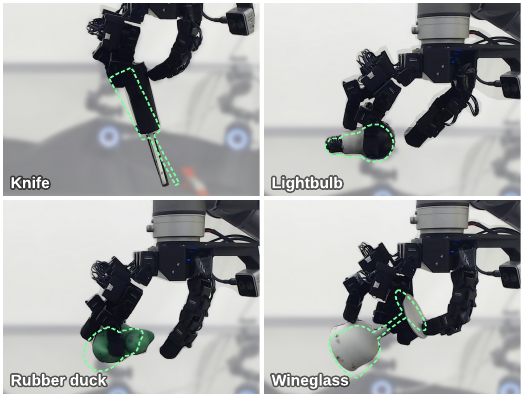}
    \caption{\textbf{Difficult out-of-distribution goals reached on the real robot} (palm down, state policy; dashed green outline: goal).}
    \label{fig:hard_goals}
\end{figure}

\begin{figure}[t]
    \centering
    \includegraphics[width=\columnwidth]{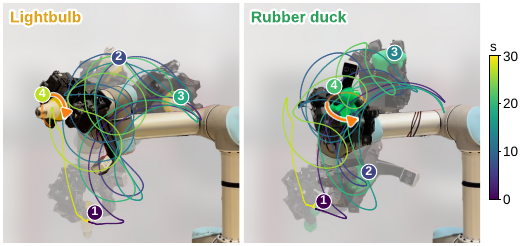}
    \caption{\textbf{Rotating object during large arm motion.} During a 30\,s arm trajectory, the depth policy keeps the lightbulb and the duck spinning. Four poses are shown, with the object path colored by time; the orange arrow is the commanded rotation.}
    \label{fig:motion_robustness}
\end{figure}

\begin{figure*}[t]
    \centering
    \includegraphics[width=\textwidth]{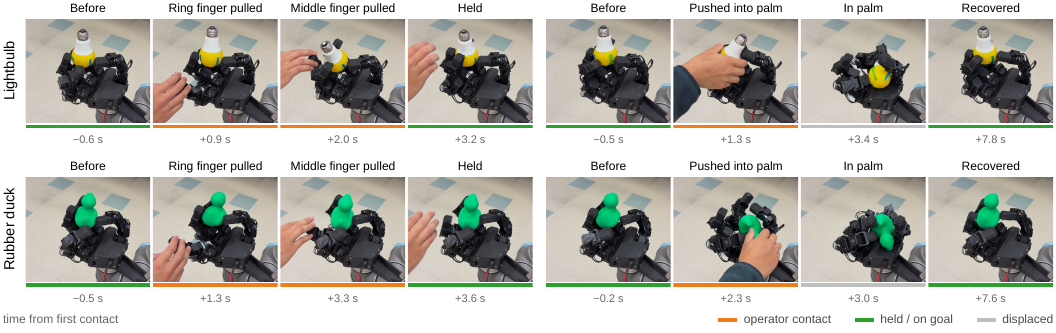}
    \caption{\textbf{Robustness to physical disturbances (depth policy, no filter).} Left: a finger is perturbed while the hand holds the object, and the hand keeps the object on goal. Right: the object is pushed into the palm, and the hand recovers and returns it to the goal about 4\,s after release.}
    \label{fig:qualitative_robustness}
\end{figure*}

\begin{figure*}[t]
    \centering
    \includegraphics[width=\textwidth]{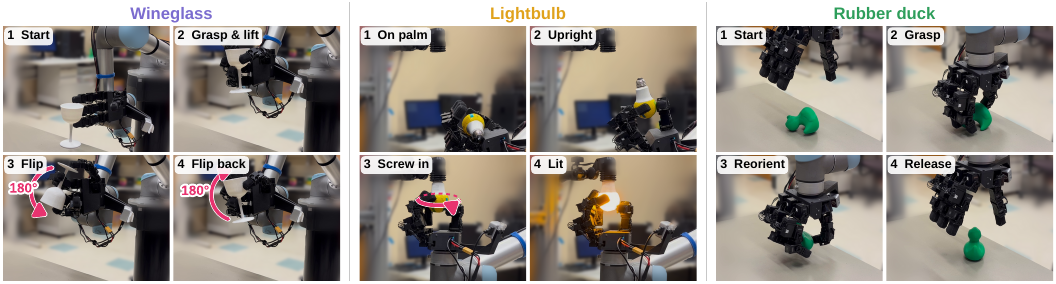}
    \caption{\textbf{Multi-step arm and hand tasks} with the depth policy, using only the wrist camera (no motion capture or external cameras).}
    \label{fig:task_sequence}
\end{figure*}

\begin{figure}[t]
    \centering
    \includegraphics[width=\columnwidth]{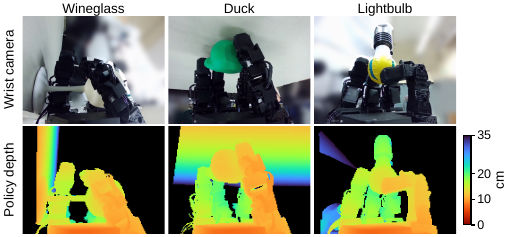}
    \caption{\textbf{Wrist-camera input during the tasks of Fig.~\ref{fig:task_sequence}}, with clutter in the background. Camera image (top) and the depth the policy receives (bottom). }
    \label{fig:depth_clutter}
\end{figure}

We first test performance, reliability and safety over long sequences of goals on the real robot. We test four configurations, the state policy and the depth policy, each with and without its residual safety filter at $w = 0.5$ (the middle ground of the simulated sweep), on each of the four objects at six wrist orientations that point gravity along each axis of the hand (Fig.~\ref{fig:wrist_orientations}). Each of these 96 runs attempts 50 consecutive goals with a 10\,s window per goal. The goals come from a held-out evaluation dataset of 100{,}000 unseen grasps, and for each object and wrist orientation, all four configurations attempt the same goal sequence. If the object is dropped or gets stuck, we reset the hand to a fixed grasp and continue with the next goal. If the hand disconnects, we reset it the same way and redo the current goal, which is not counted (criteria in Appendix~\ref{app:eval}). All runs take place inside the motion-capture cage. Motion capture measures whether goals are reached and objects are dropped, and gives the object pose input only to the state policy (Sec.~\ref{sec:hardware}). We keep the lightbulb's twist command at zero and study spinning separately (Sec.~\ref{sec:twist}). In total, the runs attempt 4{,}800 goals, 13.3\,h of goal windows on the real robot (Table~\ref{tab:real_main_col}, Fig.~\ref{fig:goal_outcomes}).

Table~\ref{tab:real_main_col} and Fig.~\ref{fig:goal_outcomes} give the results. Without the filter, the state policy reaches 95.2\,\% of all goals at the coarse success tolerance and 75.6\,\% at the precise success tolerance. Ordered by performance, the objects are lightbulb $>$ rubber duck $>$ knife $>$ wineglass. The wineglass, trained in simulation on looser success tolerances ($\tau_p = 1.5$\,cm, $\tau_R = 0.3$\,rad), reaches only 141 of 300 goals at the precise success tolerance. For each object, coarse success is similar to simulation, and drops are, if anything, rarer, although the real and simulated evaluation protocols differ slightly (Appendix~\ref{app:eval}). The depth policy matches simulation on the lightbulb and rubber duck (95\,\% and 88\,\% coarse success) but not on the knife and wineglass (54\,\% and 50\,\%, against 90\,\% and 84\,\% in simulation), and the wineglass drops often (51 of 300 goals). To see whether this gap is perceptual, we compare the depth policy's auxiliary pose estimate with motion capture, replaying the recorded stereo pairs: for the duck and lightbulb it is as accurate as in simulation, but for the knife and wineglass the real errors have long tails that simulation does not show (Appendix~\ref{app:depth_gap}), which may explain their larger drop in performance.

Goal by goal, the filter costs some performance: with the state policy, coarse success falls from 95.2\,\% to 91.5\,\% and precise success from 75.6\,\% to 67.2\,\%; with the depth policy, from 71.8\,\% to 66.7\,\% and from 38.3\,\% to 31.8\,\%. This is expected, since the filter does not see the goal and is never rewarded for reaching it. For safety and reliability, the filter gives large gains. With the state policy, drops fall from 16 to 4 of 1{,}200 goals, the mean time to the first drop rises from 334 to 465\,s, and the goals reached before the first drop from 31.8 to 42.9 out of 50. With the depth policy, drops fall from 88 to 46, the time to the first drop rises from 228 to 354\,s, and the goals reached before the first drop from 17.8 to 26.0. Excluding the wineglass, the depth policy's drops fall from 37 to 6. With the filter, the robot thus reaches more goals before a drop and runs longer without one, although it reaches slightly fewer goals overall (Figs.~\ref{fig:wineglass_filter_save} and~\ref{fig:more_filter_saves}; per-object trade-off in Fig.~\ref{fig:success_drop_trade_off}, Appendix~\ref{app:results}). The depth wineglass is the exception: its drops only fall from 51 to 40, again a sign of its visual sim-to-real gap.

Qualitatively, we observe that many drops, in simulation and on the real robot, are hard to anticipate. Drops do not happen only in visibly difficult grasps. An object in a seemingly stable grasp can suddenly slip after a barely visible change in the finger motion. Whether an action leads to a drop can thus depend on subtle details of the contact, which makes the safety problem hard. This may be one reason why the residual filter, which corrects every action a little, compares well with the HJ filters, which must predict when an action is unsafe before intervening. In practice, the knife and wineglass depth policies are still quite usable: this benchmark of random goals is difficult, and with easier goals they still produce very capable dexterous behavior.

\subsubsection{Spinning the Lightbulb at a Goal}
\label{sec:twist}
We also test spinning the lightbulb about its own axis with the depth policy (Fig.~\ref{fig:lightbulb_twist}). The bulb is rotationally symmetric, so depth cannot tell its rotation about its axis, and full 6D goals are impossible to reach from depth alone. With an axis goal and a goal twist, however, the policy can still spin the bulb about the goal axis. We run 50 consecutive goals with the palm up and 50 with the palm down, each with a 20\,s window and a random twist command of $+1$ or $-1$\,rad/s (protocol in Appendix~\ref{app:eval}). Without the filter, the policy reaches 46 and 44 goals within the coarse tolerance (palm up and down), makes at least one full turn in 28 and 12 of them, and averages 1.24 and 0.63 turns per goal (Table~\ref{tab:lightbulb_twist}). The filter costs some performance (43 and 41 goals reached, 0.74 and 0.36 turns per goal), and no run drops the bulb. Spinning palm down is much harder: at some bulb poses the hand can barely rotate the bulb, which may be as much a hardware limitation as a policy one. Even so, with or without the filter, the policy does not become less safe: when spinning is too hard, it reaches the goal axis and simply holds the bulb still.

\subsubsection{Difficult Out-of-Distribution Goals}
\label{sec:ood}
To stress-test the system, we generate a second evaluation dataset of grasps in a 1.5$\times$ larger box in front of the palm; only 34--52\,\% of the goals we ran lie inside the original canonical box (Appendix~\ref{app:grasp}), so the goals lie largely outside the training distribution (Fig.~\ref{fig:goal_diversity}). With the palm down and the state policy, coarse success falls to 28--42 of 50 goals per object (56--84\,\%), and to 19--39 with the filter (Table~\ref{tab:wide_goals}; protocol in Appendix~\ref{app:eval}). Safety, however, does not suffer: only the wineglass dropped, twice with and twice without the filter. Fig.~\ref{fig:hard_goals} shows some of the very hard goals that were reached, such as holding the knife by the very tip of its handle or the lightbulb without the thumb.

This suggests that the nominal policy already acts safely on out-of-distribution goals: its reward penalizes dropping, and even on deliberately harder goals the policy does not drop more. This agrees with our earlier observation that drops are often unexpected and do not happen only in hard-to-reach grasps (Sec.~\ref{sec:real_main}). The filter seems better suited to correcting such sudden, unexpected drops, as in the main real-robot experiment.

\subsubsection{Robustness to Unseen Arm Motion}
\label{sec:motion}
Next, we move the arm while the depth policy spins an object in hand (Fig.~\ref{fig:motion_robustness}). The motion changes the direction of gravity in the hand frame during the episode and adds inertial forces, neither of which occurs in training, so it tests whether the hand keeps spinning the object without dropping it under large disturbances. In each 30\,s trial, the three UR5 wrist joints follow sinusoids (up to $\pm$85$^\circ$, peak 2.75\,rad/s), while the lightbulb spins about its goal axis and the rubber duck's goal rotates continuously. For each object, we use two goal heights above the palm and run 10 trials per height, with and without the filter. The object stays in hand in 32 of 40 trials without the filter and 38 of 40 with it, with the largest gains at the higher, harder goals (Table~\ref{tab:motion_r2}; protocol in Appendix~\ref{app:eval}). At the high lightbulb goal, however, the filter sometimes keeps the bulb in hand but stops it from rotating (3 of 10 trials). Overall, the policy keeps spinning the objects under large disturbances it never saw in training, and the filter makes it safer.

\subsubsection{Qualitative Robustness}
\label{sec:qual}
We also find that the depth policy is robust to, and recovers from, physical disturbances it never experienced in training: when a person pulls fingers of the hand away from the object it holds, the remaining fingers keep the object near the goal, and when a person pushes the object into the palm, the policy regrasps it and returns it to the goal within a few seconds (Fig.~\ref{fig:qualitative_robustness}).

\subsubsection{Downstream Tasks without Motion Capture}
\label{sec:tasks}
We then use the depth policy as a skill in multi-step tasks, with the hand on the arm and hard-coded arm motions, using only the wrist camera: no motion capture, no external cameras and no markers on the objects (Fig.~\ref{fig:task_sequence}, Table~\ref{tab:tasks}, protocol in Appendix~\ref{app:eval}). The hand picks up a wineglass and flips it 180$^\circ$ in hand and back; regrasps a lightbulb lying in the palm, carries it under a socket and screws it in with a continuous twist goal until it lights up; and grasps a rubber duck lying on its side, reorients it and places it upright. The arm motions and the in-hand goals given to the pose-reaching policy are hard-coded. These tasks contain behaviors that the policy never saw in training: it closes into a grasp from a hard-coded pregrasp next to an object on the table (Fig.~\ref{fig:task_sequence}, panels 1 of the wineglass and rubber duck), and, when screwing in the lightbulb, once the bulb catches the socket thread, it can no longer move freely. The wineglass task succeeds in 7 of 10 trials without the filter and 6 of 10 with it, the lightbulb lights up in 4 of 5 and 3 of 5, and the duck ends upright in 7 of 10 and 3 of 10. The filter, at $w = 1.5$, does worse; the reasons are hard to tell, but picking up objects and screwing in the lightbulb are far outside what either the policy or the filter saw in training. Object placement and the scripted arm and pregrasp motions add their own failures. The tasks mainly show what a pose-reaching skill makes possible: it needs only a wrist camera and works with clutter in the background of the depth image (Fig.~\ref{fig:depth_clutter}). Future work could combine such goal-reaching skills with higher-level planners, such as VLM agents or VLA models.

\FloatBarrier
\section{CONCLUSIONS}

We present a recipe that trains a LEAP hand in simulation, without human data, to move objects to arbitrary poses in the hand at any wrist orientation, and that transfers zero-shot to the real robot with motion capture or a single wrist depth camera. On top of it, a learned residual safety filter, trained separately, greatly reduces drops, and its single dial $w$ allows easy tuning of performance and safety. Over 13.3 hours of real-robot evaluation, the state policy reaches 95.2\,\% of goals and the safety filter cuts its drops by 75\,\% (from 16 to 4), and because the depth policy needs only a wrist camera, it serves as a portable skill in downstream arm-and-hand tasks. Pose reaching thus provides both a general dexterous skill and a demanding testbed for studying performance and safety in high-dimensional systems.

\textbf{Limitations and future work.} Our policies are trained per object, and our filter is trained per object and per nominal policy. The depth policies still show a visual sim-to-real gap on the knife and wineglass.  Future work  includes reaching near 100\,\% success with zero drops, multi-object policies, closing the visual gap, and extending learned residual safety filters into a general recipe for high-dimensional systems that can filter any nominal policy. We hope this system serves as a first step and a platform to further investigate these questions.


\section*{ACKNOWLEDGMENT}
We used AI-based assistants to help edit the text and figures.

\bibliographystyle{IEEEtran}
\bibliography{myrefs}

\iffullversion
\clearpage
\section*{Appendix}
\label{app:start}
\setcounter{subsection}{0}
\renewcommand{\thesubsection}{\Alph{subsection}}

\subsection{Hardware and Depth Pipeline}
\label{app:hardware}

Each of the hand's 16 Dynamixel XC330-M288 servos runs in its firmware's position-control mode, a P-only position loop with a PWM limit (Table~\ref{tab:actuator}), on a 4\,Mbaud bus; the driver reads all 16 joints at up to 200\,Hz. We cover only the fingertips with 3M grip tape, since bare 3D-printed plastic has too little friction for fingertip grasps, and leave the rest of the hand as 3D-printed plastic, plastic servo housings and some (non-grippy) electrical tape over parts of the hand's wiring. This way the policy manipulates mostly with its fingertips, which we prefer since the exposed wiring throughout the hand is not modeled in simulation.

The UR5 arm joint readings arrive at 50\,Hz over RTDE, and the motion-capture system runs at about 100\,Hz. Each object copy carries 10--12 small (3\,mm) markers, so that the object frame is tracked even under heavy finger occlusion, and 9 markers on the hand give the hand frame. The markers form small bumps on the object's surface that we do not model in simulation, yet the policy transfers well. On the taped lightbulb used outside the motion-capture cage, we add a few colorful stripes so that its rotation is easy to see.

All computation for running the robot, policy and safety filter, and neural stereo network, runs on one desktop with one of its NVIDIA RTX 4080 SUPER GPUs. Each device (hand bus, arm, motion capture, camera and stereo network) runs in its own thread and keeps only its latest reading. Policy inference, including the safety filter, takes a median of 5.0\,ms for the state policy (5.6\,ms with the filter) and 5.7\,ms for the depth policy (7.4\,ms with the filter).

\textbf{Depth pipeline.} The camera is an Orbbec Gemini 305 and is mounted on the hand's wrist with a 3D-printed adapter. It streams its two color sensors at 640$\times$480 and 60\,fps, with an 18.0\,mm baseline, suited to short-range depth. We rectify the stereo images with the camera's factory calibration  and run Fast FoundationStereo \cite{wen2026fast} (checkpoint 23-36-37, TensorRT FP16) on it at 640$\times$480 resolution, with 8 iterations and a maximum disparity of 192\,px. We cap the pipeline at 30\,Hz. We post-process the output: invalid returns and returns outside $(0.001, 0.35)$\,m are set to zero, and the image is resized by nearest neighbor to 160$\times$120. Inside the policy, depth is normalized from $[0.001, 0.35]$\,m to $[0, 1]$. The policy samples the latest frame at 20\,Hz. The simulated camera renders depth directly at 160$\times$120 with the same calibrated intrinsics and extrinsics and goes through the same range gate and normalization, plus the augmentation of Table~\ref{tab:domain_randomization} (Fig.~\ref{fig:depth_sim}).

\subsection{System Identification}
\label{app:sysid}

\textbf{Actuator model.} We identify the XC330 with the BAM framework \cite{duclusaud2025extended, duclusaud2024bam} on a single-servo pendulum test bench (Fig.~\ref{fig:sysid}): recording 240 different trajectories of about 6\,s each, covering 10 mass/arm-length setups (0.056--0.296\,kg, 0.08--0.15\,m), 6 trajectory types and 4 firmware P gains. We fit BAM's m5 model, a voltage-driven DC-motor model with Coulomb, viscous, load-dependent and Stribeck friction. Since all 16 joints use the same servo, the parameters identified on the single servo are used for every joint (Table~\ref{tab:actuator}). We also simulate the firmware's position loop with its PWM and current limits and the supply voltage drop across the shared bus following \cite{duclusaud2024bam}. 

\textbf{Latency calibration.} We command a chirp (0.53 to 1.67\,Hz, 30\,s) on all 16 joints of the real hand, replay the same commands in a delay-free simulation, and find the time shift that best aligns the real and simulated joint trajectories. This gives a total lag of 17.5\,ms from command to joint reading. A joint reading is measured to be 4.9\,ms old when used, which leaves an actuation delay of 12.6\,ms. Aligning the depth images the same way gives 75\,ms from command to image, and subtracting the actuation delay leaves 62.4\,ms of camera pipeline lag. The randomized delays in simulation cover these measurements: 0--50\,ms for joint readings, 0--33\,ms for actuation and 0--100\,ms for depth (Tables~\ref{tab:domain_randomization} and~\ref{tab:actuator}). With the identified model and delays, simulated joints follow the real ones with a per-joint RMSE of 0.45$^\circ$, against 1.14$^\circ$ without delays (Fig.~\ref{fig:sysid_check}).

\begin{table}[t]
\centering
\scriptsize
\setlength{\tabcolsep}{2.2pt}
\renewcommand{\arraystretch}{1.05}
\caption{Identified XC330 actuator model and its randomization. $\pm x$: $\mathcal{U}(-x,x)$; drawn independently for each joint at every reset, except the armature, supply voltage and bus resistance (once per environment) and the command channel (every step).}
\label{tab:actuator}
\begin{tabular}{@{}lrll@{}}
\toprule
Parameter & Value & Unit & Randomization \\
\midrule
\multicolumn{4}{@{}l}{\textit{BAM m5 model, identified on a test bench}} \\
Torque constant $k_t$ & 0.571 & N\,m/A & $\times\mathcal{U}(0.9,1.1)$ \\
Winding resistance $R$ & 2.327 & $\Omega$ & $\times\mathcal{U}(0.9,1.1)$ \\
Armature & $1.83{\times}10^{-3}$ & kg\,m$^2$ & $\times\mathcal{U}(0.9,1.1)$ \\
Base friction & $4.85{\times}10^{-3}$ & N\,m & $\times\mathcal{U}(0.9,1.1)$ \\
Viscous friction & $3.97{\times}10^{-3}$ & N\,m\,s/rad & $\times\mathcal{U}(0.9,1.1)$ \\
Load friction, motor$^{*}$ & 0.421 & -- & $\times\mathcal{U}(0.9,1.1)$ \\
Load friction, ext.\ Stribeck$^{*}$ & 0.0997 & -- & $\times\mathcal{U}(0.9,1.1)$ \\
Stribeck velocity$^{*}$ & 0.255 & rad/s & $\times\mathcal{U}(0.9,1.1)$ \\
Stribeck exponent$^{*}$ & 1.824 & -- & $\times\mathcal{U}(0.9,1.1)$ \\
Stribeck friction$^{*}$ & $2.65{\times}10^{-9}$ & N\,m & fixed \\
Load friction, external$^{*}$ & $8.84{\times}10^{-8}$ & -- & fixed \\
Load friction, motor Stribeck$^{*}$ & $4.88{\times}10^{-7}$ & -- & fixed \\
\addlinespace[2pt]
\multicolumn{4}{@{}l}{\textit{Firmware, encoder and supply}} \\
Position P-gain & 1100 & -- & fixed \\
Error gain$^{*}$ & $2.88{\times}10^{-3}$ & duty/rad & fixed \\
PWM limit, base joints$^{a}$ & 300 & -- & fixed \\
PWM limit, other joints & 200 & -- & fixed \\
Current limit & 1.8 & A & fixed \\
Encoder bias & 0 & rad & $\pm0.03$ \\
Supply voltage $v_{\mathrm{in}}$ & 5.0 & V & $\mathcal{U}(4.5,5.25)$ \\
Bus resistance & 0 & $\Omega$ & $\mathcal{U}(0,0.4)$ \\
\addlinespace[2pt]
\multicolumn{4}{@{}l}{\textit{Command channel (physics steps at 120\,Hz)}} \\
Latency, shared by joints & -- & steps & $\mathcal{U}\{0,\dots,4\}$ \\
Dropped command & -- & -- & 3\,\%/control step \\
\quad hold after a drop & -- & steps & $\mathcal{U}\{1,\dots,24\}$ \\
\bottomrule
\end{tabular}
\vspace{3pt}
\begin{minipage}{\columnwidth}\scriptsize
$^{*}$As defined in BAM \cite{duclusaud2025extended}.\par
$^{a}$The MCP joint of each finger and the thumb's CMC joint. PWM limits are in register units (full scale 885).
\end{minipage}
\end{table}

\begin{figure}[t]
    \centering
    \includegraphics[width=\columnwidth]{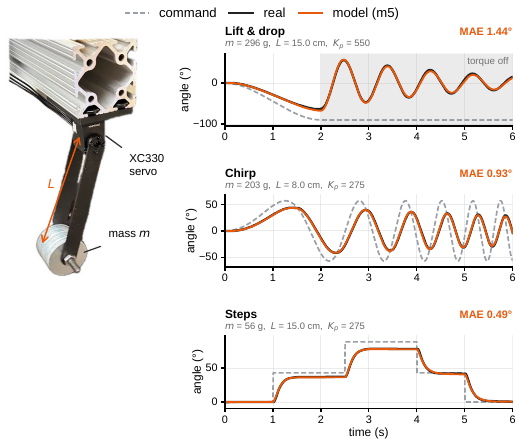}
    \caption{\textbf{Actuator identification of the XC330 servo.} Left: pendulum test bench (mass $m$, arm length $L$). Right: commanded, real and identified-model angles on three test trajectories.}
    \label{fig:sysid}
\end{figure}

\begin{figure}[t]
    \centering
    \includegraphics[width=\columnwidth]{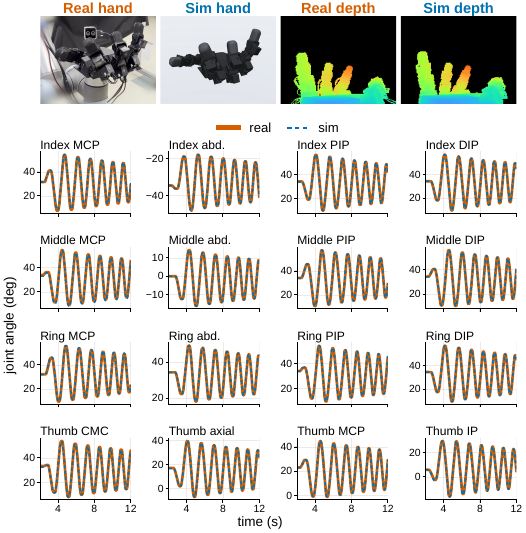}
    \caption{\textbf{Sim-to-real check of the identified hand.} The same chirp on all 16 actions, played on the real hand and replayed in simulation. Top: hand and wrist depth at the same instant. Bottom: all 16 joints, real (solid) vs.\ simulated (dashed).}
    \label{fig:sysid_check}
\end{figure}

\subsection{Simulation}
\label{app:sim}

We use mjlab 1.5 on MuJoCo Warp 3.10. The hand is fixed in free space with no floor. The wrist camera uses mjlab's camera sensor, which renders depth with MuJoCo Warp's ray tracing directly at 160$\times$120. Each control step is 6 physics steps of 1/120\,s, so the policy runs at 20\,Hz, and the actuator model (Table~\ref{tab:actuator}) runs at the physics rate. Episodes last 20\,s (400 control steps), and gravity is 9.81\,m/s$^2$. The convex decompositions have 4, 3, 5 and 32 parts for the knife, lightbulb, rubber duck and wineglass (Fig.~\ref{fig:objects}).

\begin{table}[t]
\centering
\scriptsize
\setlength{\tabcolsep}{1pt}
\renewcommand{\arraystretch}{1.05}
\caption{Per-object settings.}
\label{tab:objects}
\begin{tabular*}{\columnwidth}{@{\extracolsep{\fill}}lcccc@{}}
\toprule
 & Knife & Lightbulb & Rubber duck & Wineglass \\
\midrule
Mass, nominal [g] & 40 & 27 & 46.7 & 62 \\
Mass DR, $\log\mathcal{U}$ [g] & 20--90 & 10--50 & 20--70 & 30--90 \\
Bounding box [mm] & 39$\times$47$\times$210 & 57$\times$57$\times$103 & 59$\times$81$\times$85 & 80$\times$80$\times$137 \\
Goal$^{a}$ & pose & axis & pose & axis \\
Grasps in dataset & 451{,}681 & 453{,}131 & 439{,}212 & 421{,}949 \\
\addlinespace[2pt]
\multicolumn{5}{@{}l}{\textit{Training goal tolerance (A1 final, A2, A3, B1)}} \\
Position [cm] & 1.0 & 1.0 & 1.0 & 1.5 \\
Orientation [rad] & 0.2 & 0.12 & 0.2 & 0.3 \\
Twist [rad/s] & 0.35 & 0.35 & 0.35 & 0.35 \\
\addlinespace[2pt]
\multicolumn{5}{@{}l}{\textit{Training goal tolerance, depth stages (A4, B2)}} \\
Pos.\ / orient.\ [cm / rad] & 2.0 / 0.4 & 2.0 / 0.4 & 2.0 / 0.4 & 2.0 / 0.4 \\
\addlinespace[2pt]
\multicolumn{5}{@{}l}{\textit{Goal command}} \\
Twist rate [rad/s]$^{b}$ & 0 & $\mathcal{U}(-1.5,1.5)$ & 0 & 0 \\
Hold $h_{\mathrm{req}}$, $\mathcal{U}\{\cdot\}$ [steps]$^{c}$ & 10--20 & 10--60 & 10--20 & 10--20 \\
\addlinespace[2pt]
\multicolumn{5}{@{}l}{\textit{Training}} \\
A1 environments (2 GPUs) & 16{,}384 & 16{,}384 & 16{,}384 & 32{,}768 \\
\bottomrule
\end{tabular*}
\vspace{2pt}
\begin{minipage}{\columnwidth}\scriptsize
$^{a}$Pose: position + orientation. Axis ($z$-symmetric objects): position + direction of the object's $z$-axis, so the orientation error is a tilt angle.
$^{b}$About the goal axis; set to 0 (hold still) with probability 0.5.
$^{c}$A leaky counter: +1 per step inside all tolerances, $-1$ outside; the goal is complete when the counter reaches $h_{\mathrm{req}}$.
\end{minipage}
\end{table}

We use the implicit-fast integrator, the Newton solver with 10 iterations and 20 line-search steps, and elliptic friction cones with an impedance ratio of 10. The hand colliders take contact priority over the object, so the hand's friction coefficients determine every hand--object contact, and we randomize only the hand's friction. Only the fingertip contacts also model torsional and rolling friction, which lets a fingertip resist the object twisting under it; the other links of the hand model sliding friction only. Contact sensors between the fingertips and the object, between the other links and the object, and among the hand's own links provide the contact rewards and the fingertip-contact observation. An object's mass is distributed over its convex parts by volume.

\subsection{Domain Randomization}
\label{app:dr}
Tables~\ref{tab:domain_randomization} and~\ref{tab:actuator} list every randomized parameter and how often it is drawn: per environment, per episode or per control step. The latencies on the observations and actuation are matched to the delays measured on the real robot (Appendix~\ref{app:sysid}).

\begin{table}[t]
\centering
\scriptsize
\setlength{\tabcolsep}{2.2pt}
\renewcommand{\arraystretch}{1.05}
\caption{Domain randomization, shared by every training stage (actuator randomization in Table~\ref{tab:actuator}). $\pm x$: $\mathcal{U}(-x,x)$; drawn per environment (env), episode (ep.) or control step (step).}
\label{tab:domain_randomization}
\begin{tabular}{@{}>{\raggedright\arraybackslash}p{3.15cm}>{\raggedright\arraybackslash}p{4.35cm}l@{}}
\toprule
Quantity & Range or distribution & Drawn \\
\midrule
\multicolumn{3}{@{}l}{\textit{Contact friction (A1 curriculum start $\to$ final; final after A1)}} \\
Fingertip sliding $\mu$ & $\mathcal{U}(2.5,3.0)\to\mathcal{U}(0.3,1.2)$ & ep. \\
Non-fingertip sliding $\mu$ & $\mathcal{U}(1.0,1.5)\to\mathcal{U}(0,0.6)$ & ep. \\
Torsional $\mu$ ($\times10^{-3}$) & $\log\mathcal{U}(7,20)\to\log\mathcal{U}(0.1,20)$ & ep. \\
Rolling $\mu$ ($\times10^{-3}$) & $\log\mathcal{U}(1.7,5)\to\log\mathcal{U}(0.01,5)$ & ep. \\
\addlinespace[2pt]
\multicolumn{3}{@{}l}{\textit{Inertial$^{a}$}} \\
Robot link mass \& inertia & $\times e^{2\alpha}$, $\alpha\sim\mathcal{U}(-0.15,0.15)$ & env \\
Robot link CoM & $\pm$1\,mm per axis & env \\
Object mass & $\log\mathcal{U}$ over its range (Table~\ref{tab:objects}) & env \\
Object CoM & $\pm$10\,\% of bounding box per axis & env \\
Object inertia stretch & $\mathcal{U}(-0.05,0.05)$ & env \\
\addlinespace[2pt]
\multicolumn{3}{@{}l}{\textit{Geometry}} \\
Robot link frames & $\pm$0.3\,mm, $\pm$0.01\,rad per axis & env \\
Robot colliders & $\pm$0.5\,mm, $\pm$0.01\,rad per axis & env \\
Robot collider size & $\times\mathcal{U}(0.97,1.02)$ (boxes) & env \\
Object colliders & $\pm$0.5\,mm per axis & env \\
\addlinespace[2pt]
\multicolumn{3}{@{}l}{\textit{Perturbations and initial state}} \\
Object push force$^{b}$ & $2m\,\hat{\mathbf{u}}$\,N, $p=0.2$ new, $\times0.9$/80\,ms & step \\
Wrist orientation & uniform on $SO(3)$ & ep. \\
Initial grasp & uniform over the object's grasp set & ep. \\
\addlinespace[2pt]
\multicolumn{3}{@{}l}{\textit{Proprioception $o^{\mathrm{prop}}$ (A2--A4, B1--B2)}} \\
Joint position noise & $\pm$0.25\,\% of the joint's range & step \\
Command residual noise & $\pm$0.005\,rad & step \\
Latency$^{c}$ & $\mathcal{U}\{0,1\}$ control steps (0--50\,ms) & step \\
\addlinespace[2pt]
\multicolumn{3}{@{}l}{\textit{Object pose $x$ (A2, A3, B1)}} \\
Position noise & $\pm$3\,mm per axis & step \\
Orientation noise & $\pm$0.035\,rad per rotation-vector axis & step \\
Latency$^{c}$ & $\mathcal{U}\{0,1,2\}$ control steps (0--100\,ms) & step \\
\addlinespace[2pt]
\multicolumn{3}{@{}l}{\textit{Depth camera $I$ (A4, B2)$^{d}$}} \\
Camera position & $\pm$3\,mm per axis & ep. \\
Camera orientation & $\pm$0.03\,rad per axis & ep. \\
Field of view & $\times\mathcal{U}(0.98,1.02)$ & ep. \\
Image latency$^{c}$ & $\mathcal{U}\{0,1,2\}$ control steps (0--100\,ms) & step \\
Pixel jitter std$^{*}$ & 0.2\,px & step \\
Disparity noise std$^{*}$ & 0.025 & step \\
Surface-normal noise std$^{*}$ & 20 & step \\
Pixel dropout$^{*}$ & $p=0.001$ & step \\
Random-value pixels$^{*}$ & $p=0.001$ & step \\
Stick artifacts$^{*}$ & $p=5{\times}10^{-4}$, up to 10$\times$2\,px & step \\
Distractor objects & 6 primitives, each $p=0.6$ & ep. \\
\bottomrule
\end{tabular}
\vspace{2pt}
\begin{minipage}{\columnwidth}\scriptsize
$^{a}$Mass, inertia and CoM are perturbed jointly through the pseudo-inertia matrix \cite{rucker2022smooth}, which keeps the mass positive and the inertia tensor positive definite.
$^{b}$Following \cite{qi2022hora}: force applied at the CoM in a uniformly random direction $\hat{\mathbf{u}}$, $m$ is object mass in kg, a new push starts with probability 0.2 per step and decays $\times$0.9 every 80\,ms.
$^{c}$Redrawn every control step but kept with probability 0.8, so it behaves as a latency rather than reordering frames.
$^{d}$160$\times$120, 65.8$^\circ$ vertical FoV; returns outside [0.001, 0.35]\,m are set to 0. Distractors are visual only (slab, box, capsule, cylinder, sphere, ellipsoid) with random size, orientation, and drift.
$^{*}$Depth noise model of DextrAH-G \cite{lum2024dextrahg}, with parameters in its code's units; $p$ is per pixel.
\end{minipage}
\end{table}

\subsection{Grasp Dataset Generation}
\label{app:grasp}

\textbf{Grasp dataset.} We first use LightningGrasp \cite{yin2025lightning} to generate 100{,}000 grasps per object. LightningGrasp supports the LEAP hand, but not our version with the redesigned palm, so we add a configuration with our hand model, contacts restricted to the fingertip pads and the palm. It places the object uniformly in a canonical box in front of the palm ($x \in [0, 10]$\,cm, $y \in [-3.5, 3.5]$\,cm, $z \in [4.7, 14.5]$\,cm in the palm frame) and searches for force-balanced grasps; we refer to \cite{yin2025lightning} for the details. We then expand these seeds into a dense set of 500{,}000 grasps per object with the RRT formulation of DexterityGen \cite{yin2025dexteritygen}. We repeatedly sample a random hand configuration and object pose (joint angles uniform within the joint limits, object position uniform in the canonical box, orientation uniform over $SO(3)$), find its nearest grasp among 2048 random grasps of the current dataset under the distance $d = \lVert q - q' \rVert_2 + 50\,\lVert p - p' \rVert_2 + 3\,\theta(R, R')$ over joint angles $q$, object position $p$ (in m) and object rotation $R$, with $\theta$ the geodesic angle, and step 20\,\% of the way from that grasp toward the sample (linearly in joint angles and position, by slerp in rotation). The parent's contact points move with the object, the fingers are re-solved by inverse kinematics, and the new grasp is added to the dataset if it passes the generator's collision check. We finally filter every dataset against our simulator's own geometry, because the grasp generator's collision check missed some penetrating grasps. This removes 9.4--15.6\,\% of the grasps, and leaves 422k--453k per object (Table~\ref{tab:objects}, Fig.~\ref{fig:grasp_dataset}). Most goals are therefore likely reachable, but they are not verified in simulation, so some may be physically infeasible.

\begin{figure}[t]
    \centering
    \includegraphics[width=\columnwidth]{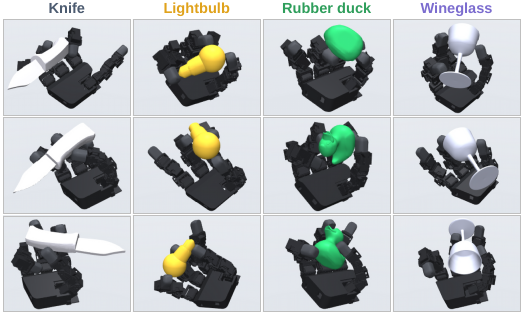}
    \caption{Example grasps from the grasp datasets generated with LightningGrasp \cite{yin2025lightning} and expanded by RRT.}
    \label{fig:grasp_dataset}
\end{figure}

\subsection{Task Environment}
\label{app:task}

\textbf{Goal completion in training.} The required hold $h_{\mathrm{req}}$ is drawn per goal from $\mathcal{U}\{10, \dots, h_{\max}\}$ steps (0.5--1\,s, and up to 3\,s for the lightbulb). 
The objects differ only in the settings listed in Table~\ref{tab:objects}: mass, goal type (pose or axis), training goal tolerances, twist range and hold range.

\textbf{Goal selection.} With the goal curriculum early in training, the next goal is not drawn uniformly from the dataset but near the current state, following DexterityGen \cite{yin2025dexteritygen}; the goal curriculum makes goals more random as the policy improves (Appendix~\ref{app:curric}).

\subsubsection{Resets and Terminations}
\label{app:resets}

The reset grasp is drawn uniformly from the grasp dataset, and the wrist orientation uniformly on $SO(3)$. Reaching the time limit is treated as a time-out rather than a failure: PPO bootstraps the value of the last state, so the policy is not penalized for an episode running out. An episode terminates when the object is more than 0.5\,m from the hand root (a drop), which also cuts off all future reward, and a NaN guard resets an environment if the simulation diverges. The lengths of the first episodes are staggered so that resets are spread out in time across the parallel environments.

\subsubsection{Action Space}
\label{app:action}
Our action is an integrator: each step adds a bounded increment to the previous position target, $q^{\mathrm{cmd}}_t = q^{\mathrm{cmd}}_{t-1} + \Delta_t$ (Eq.~\eqref{eq:action}). A study of action spaces for robot learning \cite{aljalbout2024role} finds that setting the target relative to the measured position, $q^{\mathrm{cmd}}_t = q_t + a_t$, tracks better. The integrator also makes the previous target $q^{\mathrm{cmd}}_{t-1}$ hidden state, which we expose to the policy by including the command residual $q^{\mathrm{cmd}} - q$ in its observations (Sec.~\ref{sec:obs}). We nevertheless found the integrator more reliable for system identification. The commanded target is a deterministic function of the actions alone, so the same command sequence produces the same targets on the real hardware and in simulation. If the target were built from the measured position, it would depend on which reading the real robot happened to use, how stale that reading was, and when it was read. Fig.~\ref{fig:sysid_check} shows that, with this action, simulation and the real hand track the same chirp commands closely. Without the 0.3\,rad clip, a finger that stalls (e.g., when blocked by a collision) would wind up its target and overshoot when released.

Inspired by TeleDexter \cite{li2026towards}, we also randomly drop joint commands in simulation to mimic dropped packets on the shared servo bus. In every control step, each joint independently misses its new command with probability 3\,\% and holds its last received target for $\mathcal{U}\{1,\dots,24\}$ physics steps (8--200\,ms).

In addition, a randomized command latency of $\mathcal{U}\{0,\dots,4\}$ physics steps (0--33\,ms), shared by all joints, is applied before the actuator model (Table~\ref{tab:actuator}).

\subsubsection{Observations}

Table~\ref{tab:observations} lists every observation. On the real robot, the gravity direction is computed from the arm's end-effector orientation, and joint positions are read from the servo encoders. The privileged observations add short histories of fingertip and object poses and fingertip contact forces, and the privileged goal $g^{\mathrm{priv}}$ adds the goal relative to the object. The critic alone also receives the fingertip friction coefficients and encoder offsets sampled by domain randomization for each episode, and the goal-completion state: the errors $e_p$ and $e_R$, whether the object is within the tolerances, the hold progress, the required hold length and the twist error.

\begin{table}[t]
\centering
\scriptsize
\setlength{\tabcolsep}{2.2pt}
\renewcommand{\arraystretch}{1.05}
\caption{Observations, grouped by the symbols of Fig.~\ref{fig:method}, all in the hand frame. Dim: per-frame size; Hist.: frames stacked.}
\label{tab:observations}
\begin{tabular}{@{}>{\raggedright\arraybackslash}p{6.3cm}rc@{}}
\toprule
Term & Dim & Hist. \\
\midrule
\multicolumn{1}{@{}l}{\textit{$o^{\mathrm{prop}}$ proprioception}} & \textit{} & \\
\quad Projected gravity & 3 & 1 \\
\quad Last smoothed action $s_{t-1}$ & 16 & 1 \\
\quad Command residual $(q^{\mathrm{cmd}}{-}q)/0.1$ & 16 & 1 \\
\quad Joint position$^{a}$ & 16 & 10 \\
\addlinespace[2pt]
\multicolumn{1}{@{}l}{\textit{$x$ object pose (motion capture)}} & \textit{} & \\
\quad Object position & 3 & 1 \\
\quad Object orientation (quaternion) & 4 & 1 \\
\addlinespace[2pt]
\multicolumn{1}{@{}l}{\textit{Goal (no ResSafe network reads any goal term)}} & \textit{} & \\
\quad\textit{$g$ goal} & & \\
\qquad Goal pose in hand frame$^{b}$ & 7 & 1 \\
\quad\textit{$g^{\mathrm{priv}}$ privileged goal} & & \\
\qquad $g$ & 7 & -- \\
\qquad Goal relative to object & 7 & 1 \\
\quad\textit{$g^{\mathrm{crit}}$ critic goal} & & \\
\qquad $g^{\mathrm{priv}}$ & 14 & -- \\
\qquad Position / orientation error & 2 & 1 \\
\qquad In-gate flag & 1 & 1 \\
\qquad Hold progress & 1 & 1 \\
\qquad Required hold length & 1 & 1 \\
\qquad Twist error & 3 & 1 \\
\addlinespace[2pt]
\multicolumn{1}{@{}l}{\textit{$I$ wrist depth}} & \textit{} & \\
\quad Depth image (Table~\ref{tab:domain_randomization}) & 160$\times$120 & 1 \\
\addlinespace[2pt]
\multicolumn{1}{@{}l}{\textit{$o^{\mathrm{priv}}$ privileged state}} & \textit{} & \\
\quad Projected gravity & 3 & 1 \\
\quad Last smoothed action $s_{t-1}$ & 16 & 1 \\
\quad Command residual & 16 & 1 \\
\quad Joint position (true angle) & 16 & 1 \\
\quad Joint velocity & 16 & 1 \\
\quad Fingertip position (4 tips, $xyz$) & 4$\times$3 & 5 \\
\quad Fingertip velocity & 4$\times$3 & 1 \\
\quad Object position & 3 & 5 \\
\quad Object orientation (quaternion) & 4 & 5 \\
\quad Object linear velocity & 3 & 1 \\
\quad Object angular velocity & 3 & 1 \\
\quad Fingertip contact (touch, $\log(1{+}|f|)$) & 4$\times$2 & 1 \\
\addlinespace[2pt]
\multicolumn{1}{@{}l}{\textit{$o^{\mathrm{crit}}$ critic}} & \textit{} & \\
\quad $o^{\mathrm{priv}}$ & 188 & -- \\
\quad Fingertip friction draw & 4 & 1 \\
\quad Encoder-bias draw & 16 & 1 \\
\addlinespace[2pt]
\multicolumn{1}{@{}l}{\textit{ResSafe inputs}} & \textit{} & \\
\quad $a^{\mathrm{nom}}$ nominal action$^{c}$ & 16 & 1 \\
\quad $w$ correction penalty & 1 & 1 \\
\midrule
\multicolumn{3}{@{}l}{\textit{Network inputs (vector width; $I$ enters through the depth encoder)}} \\
\quad $\pi^{\mathrm{priv}}$ (A1): $o^{\mathrm{priv}}, g^{\mathrm{priv}}$ & 202 & \\
\quad $V$ (A1, A3): $o^{\mathrm{crit}}, g^{\mathrm{crit}}$ & 230 & \\
\quad $\pi^{\mathrm{s}}$ (A2, A3): $o^{\mathrm{prop}}, x, g$ & 209 & \\
\quad $\pi^{\mathrm{d}}$ (A4): $o^{\mathrm{prop}}, g, I$ & 202 & \\
\quad $\pi_\delta$ (B1): $o^{\mathrm{prop}}, x, w, a^{\mathrm{nom}}$ & 219 & \\
\quad $V_\delta$ (B1): $o^{\mathrm{crit}}, w, a^{\mathrm{nom}}$ & 225 & \\
\quad $\pi_\delta^{\mathrm{d}}$ (B2): $o^{\mathrm{prop}}, I, w, a^{\mathrm{nom}}$ & 212 & \\
\bottomrule
\end{tabular}
\vspace{2pt}
\begin{minipage}{\columnwidth}\scriptsize
$^{a}$Limit-normalized to $[-1,1]$ and read with the encoder bias; 10 frames $=$ 0.5\,s.
$^{b}$Knife, rubber duck: position + quaternion. Lightbulb, wineglass ($z$-symmetric): position + goal $z$-axis + commanded twist rate.
$^{c}$In training, perturbed by $s\boldsymbol{\epsilon}$, $s\sim\mathcal{U}(0,0.5)$.
\end{minipage}
\end{table}

\begin{figure}[t]
    \centering
    \includegraphics[width=\columnwidth]{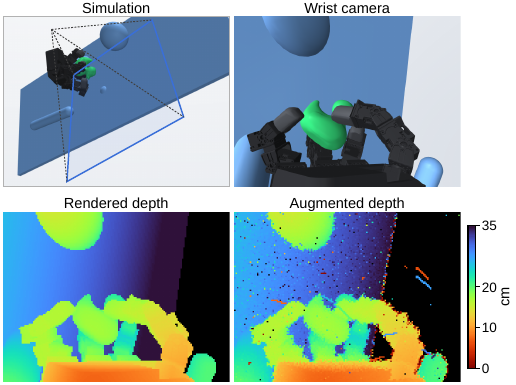}
    \caption{\textbf{Simulated depth observations.} Scene with distractors, wrist camera view, rendered depth and augmented depth.}
    \label{fig:depth_sim}
\end{figure}

\subsubsection{Reward}
\label{app:reward}

When a goal is complete and a new one is drawn, the potentials are reset to the new goal's current values (the hold counter and its potential return to zero, and the position, orientation and twist terms give zero reward on that one step instead of rewarding the jump in error to the new goal), so completing a goal never registers as a loss of potential.

\subsection{Curricula}
\label{app:curric}

Each curriculum has a progress $p \in [0, 1]$ that adapts to the policy's completion rate, as in the fine-tuning curriculum of the OmniReset code release \cite{yin2026emergent}. We keep a buffer of the outcomes of the last 500 goals across all environments (1 if the goal was completed, 0 if the episode ended first). Every 10 control steps, each curriculum moves its progress up by $1/2048$ if the fraction of completed goals is above its threshold, and down by $1/2048$ if it is below a slightly lower threshold (0.50/0.49 for the goal curriculum and 0.55/0.54 for the friction curriculum). The progress then sets the parameters of each curriculum (Fig.~\ref{fig:curriculum_details}); Fig.~\ref{fig:curriculum_progress} shows the progress during A1.
For the goal curriculum we build on the nearest-neighbor goal rule of DexterityGen \cite{yin2025dexteritygen}. We draw a random subset of $N$ grasps from the dataset, rank them by their distance to the current state,
\begin{equation}
d_i = \lVert q - q_i \rVert_2 + 50\,\lVert p - p_i \rVert_2 + 2\,\theta(R, R_i),
\end{equation}
where $q$ are the measured joint angles, $(p, R)$ is the current object pose, $(q_i, p_i, R_i)$ is grasp $i$ and $\theta$ is the geodesic angle, and pick one of the 10 nearest uniformly. Goals are thus nearby and reachable at first. We shrink the random subset geometrically from 2048 to 1 as $p$ goes from 0 to 1, $N = \mathrm{round}(2048^{1-p})$. With fewer than 10 candidates, the goal is drawn uniformly from the whole dataset. The same progress also tightens the success tolerances linearly from 2\,cm, 0.4\,rad and 1.5\,rad/s to each object's final values (Table~\ref{tab:objects}), and raises the maximum required hold $h_{\max}$ from 10 to 20 steps (60 for the lightbulb).

mjlab can modify the friction of any collision geometry directly on the GPU (in other frameworks such as Isaac Lab PhysX \cite{mittal2025isaaclab}, this currently goes through the CPU). Friction is domain randomized: at every reset, the friction of each hand geometry is drawn uniformly from a range. At $p = 0$ the ranges are unrealistically high: $\mathcal{U}(2.5, 3.0)$ sliding friction for the fingertips and $\mathcal{U}(1.0, 1.5)$ for the rest of the hand. As $p$ increases, the bounds move linearly to the final ranges $\mathcal{U}(0.3, 1.2)$ and $\mathcal{U}(0.0, 0.6)$ at $p = 1$, and the torsional and rolling friction ranges widen geometrically toward zero (Table~\ref{tab:domain_randomization}). The intuition is that people also learn fine manipulation with grippy tools or helpers that make things less slippery and easier to hold.

All later stages (A2--A4, B1--B2) run with both curricula pinned at $p = 1$.

\begin{figure*}[t]
    \centering
    \includegraphics[width=\textwidth]{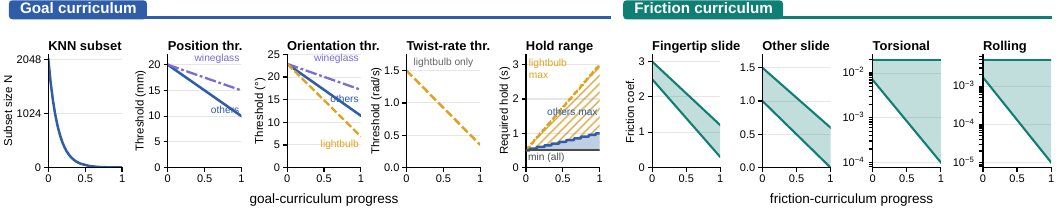}
    \caption{\textbf{Curriculum schedules} as a function of progress $p$. Left: goal curriculum (candidate subset, tolerances, required hold). Right: friction curriculum (sampled range per coefficient).}
    \label{fig:curriculum_details}
\end{figure*}

\begin{figure}[t]
    \centering
    \includegraphics[width=\columnwidth]{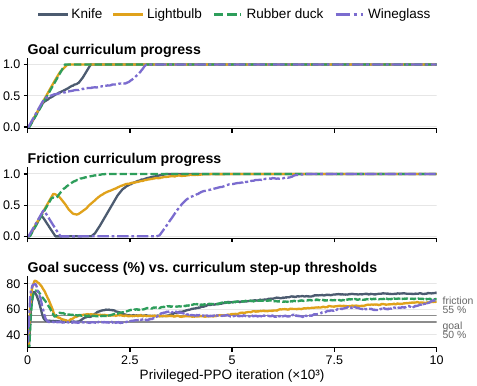}
    \caption{\textbf{Adaptive curriculum progress during A1}: goal (top) and friction (middle) curriculum progress, and the rolling goal success that drives them (bottom).}
    \label{fig:curriculum_progress}
\end{figure}

\subsection{Training Details and Compute}
\label{app:training}

Tables~\ref{tab:hp_ppo} and~\ref{tab:hp_dagger} list the hyperparameters of every stage, and Fig.~\ref{fig:training_stages} shows the training curves of the four task stages. A1 uses the rsl\_rl implementation of PPO \cite{schwarke2025rslrl}. In A2, the student executes its own actions and regresses the teacher's mean action. A3's actor is initialized from the A2 student and its critic is copied from A1, so PPO fine-tuning starts from a trained value function. In A4, the depth image is encoded by a CNN with a spatial softmax, and the policy MLP uses LayerNorm, which we found stabilized training.

\textbf{Training time.} Our runs were spread over several machines (RTX 4090, RTX 4080 SUPER and RTX A6000 GPUs), some of them shared with other jobs, so their logged durations are not comparable. Instead, the training times we report are each stage's maximum iterations multiplied by the median time per iteration (over 22 iterations of a 25-iteration run, after 3 warm-up iterations), measured for every stage and object on one machine with RTX 5090 GPUs and run as in training: A1 on two GPUs and every other stage on one GPU. A1 takes about 15--17\,h (28\,h for the wineglass since it uses twice the number of environments), A2 0.6\,h, A3 2.5--2.8\,h, A4 10\,h, B1 2.5\,h and B2 1.8\,h. A complete policy and filter for one object thus takes about 32--35\,h (46\,h for the wineglass) on this hardware. 

\textbf{Seeds.} Each deployed task policy comes from a single training run (A1--A4) per object. The ablations give their number of seeds in the captions: three for the knife curriculum and privileged-actor ablations and the rubber-duck ResSafe ablation, and one for the four-object curriculum ablation. For ResSafe, we train three B1 seeds per object and select the one with the highest goal success at each object's training goal tolerance (Table~\ref{tab:objects}), averaged over $w \in \{0, 0.5, 1, 1.5, 2\}$ in simulation. The B2 depth filter is distilled from that seed, and these selected filters are the ones deployed on the real robot and reported in every table. The HJ filters are selected from their three seeds by the same rule, averaged over the swept $\alpha$ and $\beta$.
\label{app:seeds}

\begin{table}[t]
\centering
\scriptsize
\setlength{\tabcolsep}{3pt}
\renewcommand{\arraystretch}{1.05}
\caption{PPO hyperparameters (A1, A3, B1). $\mathcal{M}_5=[1024,512,512,256,256]$ hidden units, ELU.}
\label{tab:hp_ppo}
\begin{tabular*}{\columnwidth}{@{\extracolsep{\fill}}lccc@{}}
\toprule
 & A1 & A3 & B1 \\
\midrule
Environments & 2$\times$8192$^{a}$ & 8192 & 8192 \\
Max.\ iterations$^{b}$ & 30k & 5k & 4k \\
Wall-clock$^{e}$ & 15--17\,h & 2.5--2.8\,h & 2.5--2.6\,h \\
Steps per env.\ per iteration & 24 & 24 & 24 \\
Epochs / mini-batches & 5 / 4 & 5 / 4 & 5 / 4 \\
Learning rate$^{c}$ & $5{\times}10^{-4}$ & $5{\times}10^{-4}$ & $5{\times}10^{-4}$ \\
Max.\ gradient norm & 1.0 & 1.0 & 1.0 \\
$\gamma$ / GAE $\lambda$ & 0.99 / 0.95 & 0.99 / 0.95 & 0.99 / 0.95 \\
Clip / value-loss coef. & 0.2 / 1.0 & 0.2 / 1.0 & 0.2 / 1.0 \\
Entropy coefficient & 0.002 & 0.001 & 0.0005 \\
Initial action std$^{d}$ & 1.0 & A1's & 0.4 \\
Actor / critic MLP & $\mathcal{M}_5$ / $\mathcal{M}_5$ & $\mathcal{M}_5$ / $\mathcal{M}_5$ & $\mathcal{M}_5$ / $\mathcal{M}_5$ \\
\bottomrule
\end{tabular*}
\vspace{2pt}
\begin{minipage}{\columnwidth}\scriptsize
$^{a}$Two GPUs; wineglass 2$\times$16384.
$^{b}$Deployed checkpoints (knife / lightbulb / rubber duck / wineglass): A1 28000 / 28500 / 24500 / 26500, A3 5000 / 4500 / 3500 / 5000, B1 4000.
$^{c}$Adaptive: scaled to keep the KL divergence near 0.01.
$^{d}$A3 starts from the value in the A1 checkpoint that A2 distilled (mean 0.38--0.50); its actor weights come from A2, its critic from A1.
$^{e}$Max.\ iterations at the median time per iteration, measured on RTX 5090 GPUs (A1 on two, A3 and B1 on one): A1 1.8--2.0\,s (wineglass 3.4\,s at 2$\times$16384, 28\,h), A3 1.8--2.1\,s, B1 2.2--2.4\,s, by object.
\end{minipage}
\end{table}

\begin{table}[t]
\centering
\scriptsize
\setlength{\tabcolsep}{3pt}
\renewcommand{\arraystretch}{1.05}
\caption{DAgger hyperparameters (A2, A4, B2). $\mathcal{M}_5=[1024,512,512,256,256]$, $\mathcal{M}_3=[1024,512,256]$ hidden units, ELU.}
\label{tab:hp_dagger}
\begin{tabular*}{\columnwidth}{@{\extracolsep{\fill}}lccc@{}}
\toprule
 & A2 & A4 & B2 \\
\midrule
Environments & 4096 & 512 & 512 \\
Iterations$^{a}$ & 2000 & 20000 & 3000 \\
Wall-clock$^{b}$ & 0.6\,h & 10\,h & 1.8\,h \\
Steps per env.\ per iteration & 24 & 24 & 24 \\
Epochs / mini-batches & 5 / 4 & 5 / 4 & 5 / 4 \\
Learning rate & $5{\times}10^{-4}$ & $5{\times}10^{-4}$ & $5{\times}10^{-4}$ \\
Max.\ gradient norm & 1.0 & 1.0 & 1.0 \\
Student MLP & $\mathcal{M}_5$ & $\mathcal{M}_3$ + LayerNorm & $\mathcal{M}_3$ \\
Depth encoder & -- & CNN$^{c}$ & CNN$^{c}$ \\
Aux.\ loss weight & -- & 1.0 & 1.0 \\
\bottomrule
\end{tabular*}
\vspace{2pt}
\begin{minipage}{\columnwidth}\scriptsize
$^{a}$Deployed checkpoint; A2 and A4 ran without a fixed budget.
$^{b}$Iterations at the median time per iteration, measured on one RTX 5090: A2 1.0--1.1\,s, A4 1.8\,s, B2 2.1--2.2\,s, by object.
$^{c}$Conv 32@8$\times$8/4, 64@4$\times$4/2, 64@3$\times$3/1 (ReLU); spatial-softmax keypoints plus channel means (192-D), concatenated with the vector input.
\end{minipage}
\end{table}

\begin{figure}[t]
    \centering
    \includegraphics[width=\columnwidth]{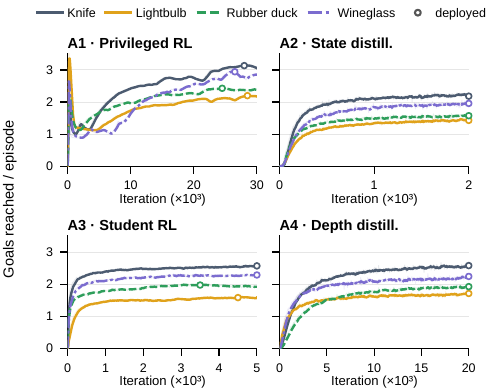}
    \caption{\textbf{Training curves of the four task stages} (goals per episode). Hollow circles: deployed checkpoints.}
    \label{fig:training_stages}
\end{figure}

\subsection{Safety Filters}
\label{app:safety}
\label{app:hj}

\subsubsection{Training the Residual Filter}
\label{app:ressafe_train}

The correction weight $w \sim \mathcal{U}(0, 2)$ is drawn per environment and episode. Unlike the task policy's A1 actor, the filter's actor reads only deployable observations ($o^{\mathrm{prop}}$, $x$, $a^{\mathrm{nom}}$ and $w$, i.e., A2's input without the goal) and can run on the robot directly; the critic receives the privileged state $o^{\mathrm{crit}}$ plus $a^{\mathrm{nom}}$ and $w$.
We train three B1 seeds per object and deploy the one selected as described in Appendix~\ref{app:seeds}; the B2 depth filter is distilled from that seed.

\subsubsection{Hamilton--Jacobi Reachability}
\label{sec:hj_reachability}
Hamilton--Jacobi (HJ) reachability characterizes the largest set of states from which a system can be kept safe for all time \cite{bansal2017hamilton}. Let $\ell : \mathcal{X} \to \mathbb{R}$ be a margin function over the state space $\mathcal{X}$, whose negative set $\{x : \ell(x) < 0\}$ defines the failure set $\mathcal{F}$. The safety value function captures the worst-case margin along a system trajectory $\xi_x^u(t)$ starting at state $x$ under control $u(\cdot)$:
\begin{equation}
V(x) = \sup_{u(\cdot)} \inf_{t \ge 0} \ell\bigl(\xi_x^u(t)\bigr).
\end{equation}
Thus $V(\cdot)$ characterizes the maximal safe set, defined as $\{x \in \mathcal{X} \mid V(x) \geq 0\}$. In our setting, $\ell$ is the drop margin of \eqref{eq:margin}.

In discrete-time, this safety value function can be obtained by solving the discrete-time Hamilton--Jacobi fixed-point equation via dynamic programming:
\begin{equation}
V(x) = \min\Bigl\{ \ell(x), \; \max_{u \in \mathcal{U}} V\bigl(f(x, u)\bigr) \Bigr\},
\end{equation}
where $f(x, u)$ represents the system dynamics. This way the safety of a state is defined as the minimum between the current safety at the state and its future safety under the best control.

\citet{fisac2019bridging} showed that this safety value function can be approximated using Q-learning and derived the following time-discounted state-action HJ fixed-point equation:
\begin{equation}
Q(x, u) = (1 - \gamma) \ell(x) + \gamma \min\Bigl\{ \ell(x), \; \max_{u' \in \mathcal{U}} Q\bigl(f(x, u), u'\bigr) \Bigr\},
\label{eq:safety-bellman}
\end{equation}
where $\gamma \in [0, 1)$ is a discount factor. In this formulation, the state-action value $Q$ and the state value $V$ are related via the optimal safety fallback policy $\pi_{\text{safe}}(x)$: 
\begin{equation}
\begin{aligned}
\pi_{\text{safe}}(x) &= \arg\max_{u \in \mathcal{U}} Q(x, u), \\
V(x) &= \max_{u \in \mathcal{U}} Q(x, u) = Q\bigl(x, \pi_{\text{safe}}(x)\bigr).
\end{aligned}
\end{equation}

By leveraging the learned $Q(x, u)$, the safe fallback policy $\pi_{\text{safe}}(x)$, and a given nominal task policy $\pi_{\text{task}}(x)$, one can construct a least-restrictive safety filter (LRSF). When a safety violation is imminent, the filter abruptly intervenes by running the safe fallback action:
\begin{equation}
    u(x) = \begin{cases}
        \pi_{\text{task}}(x), & \text{if } Q\bigl(x, \pi_{\text{task}}(x)\bigr) > 0, \\
        \pi_{\text{safe}}(x), & \text{otherwise}.
    \end{cases}
    \label{eq:lrsf}
\end{equation}

\subsubsection{Smooth Safety Filtering}
\label{sec:cbf}

An issue with LRSF is that its hard switching mechanism can result in sudden, undesirable control interventions. One solution to mitigate this is the use of control barrier functions (CBFs), which admit a smoother control law. Following \cite{oh2025safety, nakamura2026how}, the safety value function inherently satisfies the conditions of a discrete-time control barrier function (DCBF) \cite{agrawal2017discrete}. 

Leveraging this property, the Human-Centered Safety Filter (HCSF) \cite{oh2025safety} and its latent-space counterpart \cite{nakamura2026how} avoid hard switches by minimally steering the nominal policy $\pi_{\text{task}}(x)$ to ensure safety. At runtime, they solve the following optimization problem:
\begin{align}
    u(x) = \arg\min_{u \in \mathcal{U}} \quad & \|u - \pi_{\text{task}}(x)\|^2 \label{eq:hcsf} \\
    \notag \text{s.t.} \quad & Q(x, u) \geq \beta\,  Q\bigl(x, \pi_{\text{safe}}(x)\bigr),
\end{align}
where $\beta \in [0, 1)$, as in \cite{oh2025safety}, dictates how aggressively the filter overrides the nominal policy as the system approaches the safety boundary.

\subsubsection{Implementation}
\label{app:hj_impl}
In FlashSAC \cite{kim2026flashsac}, we replace the Bellman target with the safety backup \eqref{eq:safety-bellman}, $y = (1-\gamma)\,\ell(x) + \gamma \min\{\ell(x), Q'\}$, where $Q'$ is the target critic evaluated at the next state and the fallback actor's deterministic action there, with $\gamma = 0.997$ and 3-step returns. The critic outputs a distribution over 101 atoms on $[-1, 1]$, and $\mathrm{CVaR}_\alpha$ is the mean of its worst $\alpha$ fraction, taking the minimum over the two critics. The fallback actor still maximizes the mean. LRSF switches to the fallback action when $\mathrm{CVaR}_\alpha(Q(x, u_{\mathrm{task}})) \leq \epsilon$, with $\epsilon = 10^{-3}$. HCSF approximates \eqref{eq:hcsf} with a simple zeroth-order search: it evaluates 16 candidates on the segment from the nominal action to the fallback action and executes the one closest to the nominal action that satisfies $\mathrm{CVaR}_\alpha(Q(x, u)) - \epsilon \geq \beta\,(\mathrm{CVaR}_\alpha(Q(x, \pi_{\text{safe}}(x))) - \epsilon)$, the constraint of \eqref{eq:hcsf} on CVaR with the same margin $\epsilon$. We train the HJ filter on the rubber duck in the same environment as ResSafe, except that its episodes do not terminate when $\ell(x) < 0$, with 1{,}024 environments for 50k iterations (actor MLP 2$\times$128, critic 3$\times$512).

Motivated by \cite{nakamura2026how}, which mixes nominal-policy and safety-policy rollouts in training to reduce the mismatch with the state distribution at deployment, we fill the replay buffer by running the filter itself, so that training sees the same mix of nominal and filtered actions as evaluation. Half of the environments use LRSF and half HCSF, with $\alpha$ and $\beta$ drawn per episode from $\mathcal{U}(0,1)$, and the nominal action is perturbed with the same noise scale as for ResSafe, $s \sim \mathcal{U}(0, 0.5)$.

Both filters observe the same deployable, goal-free observation at run time. Only ResSafe's critic receives privileged state, since ResSafe runs only its actor at deployment, whereas the HJ filter needs its critic at run time. At evaluation, we sweep LRSF over $\alpha \in \{0.05, 0.1, 0.2, 0.3, 0.5\}$ and HCSF over $\alpha \in \{0.1, 0.2, 0.3\}$ and $\beta \in \{0.5, 0.75, 0.9, 0.95, 0.99\}$. Table~\ref{tab:filter_baselines_sweep} lists every swept setting.

\begin{table}[t]
\centering
\scriptsize
\setlength{\tabcolsep}{3pt}
\caption{Every swept setting of the HJ filter (simulation, rubber duck, state policy). TTD: time to the first drop or end (s); Corr.: average action change (\%).}
\label{tab:filter_baselines_sweep}
\begin{tabular}{@{}llc ccccc@{}}
\toprule
Gate & $\alpha$ & $\beta$ & coarse & precise & Drops & TTD & Corr. \\
\midrule
\multicolumn{3}{l}{None} & 92.1 & 88.4 & 9.1 & 154 & -- \\
\multicolumn{3}{l}{ResSafe, $w{=}0.5$} & 91.3 & 80.1 & 4.5 & 266 & 15.3 \\
\midrule
LRSF & $0.05$ & -- & 60.6 & 50.4 & 5.7 & 195 & 30.2 \\
LRSF & $0.1$ & -- & 79.3 & 71.3 & 8.2 & 154 & 16.5 \\
LRSF & $0.2$ & -- & 86.1 & 81.0 & 9.4 & 142 & 9.1 \\
LRSF & $0.3$ & -- & 87.8 & 83.4 & 9.3 & 147 & 6.7 \\
LRSF & $0.5$ & -- & 90.8 & 87.0 & 9.4 & 147 & 2.9 \\
\midrule
HCSF & $0.1$ & $0.5$ & 82.1 & 73.5 & 7.6 & 169 & 15.6 \\
HCSF & $0.1$ & $0.75$ & 80.2 & 69.8 & 5.9 & 191 & 18.3 \\
HCSF & $0.1$ & $0.9$ & 70.5 & 57.2 & 3.8 & 260 & 25.7 \\
HCSF & $0.1$ & $0.95$ & 54.5 & 40.1 & 2.3 & 310 & 33.9 \\
HCSF & $0.1$ & $0.99$ & 9.6 & 5.0 & 0.5 & 445 & 55.6 \\
HCSF & $0.2$ & $0.5$ & 87.0 & 81.7 & 8.8 & 146 & 8.7 \\
HCSF & $0.2$ & $0.75$ & 87.1 & 80.9 & 7.7 & 158 & 9.8 \\
HCSF & $0.2$ & $0.9$ & 84.7 & 75.3 & 6.0 & 201 & 14.4 \\
HCSF & $0.2$ & $0.95$ & 73.6 & 60.4 & 3.9 & 256 & 23.4 \\
HCSF & $0.2$ & $0.99$ & 21.2 & 12.4 & 0.9 & 407 & 48.5 \\
HCSF & $0.3$ & $0.5$ & 89.1 & 84.6 & 8.9 & 144 & 5.7 \\
HCSF & $0.3$ & $0.75$ & 89.6 & 84.8 & 8.1 & 156 & 6.2 \\
HCSF & $0.3$ & $0.9$ & 87.4 & 80.4 & 6.6 & 185 & 10.8 \\
HCSF & $0.3$ & $0.95$ & 82.0 & 70.8 & 4.8 & 230 & 17.2 \\
HCSF & $0.3$ & $0.99$ & 28.5 & 17.8 & 1.1 & 389 & 44.3 \\
\bottomrule
\end{tabular}
\end{table}

\subsection{Evaluation Protocols}
\label{app:eval}

\textbf{Simulation protocol.} The simulation benchmark (Sec.~\ref{sec:protocol}) runs in the same environment as training, with the same domain randomization and with both curricula at full difficulty. Every environment starts from the same fixed grasp, chosen per object, at its own random wrist orientation, and draws a new random goal from the grasp dataset for each 10\,s window. After a drop, the environment is reset to the fixed grasp at the start of the next goal window.

\textbf{Main experiment.} The evaluation goals for the real experiments come from a separate grasp dataset of 100{,}000 grasps per object. It is generated with the same procedure as the training dataset, but independently: 20{,}000 new LightningGrasp seeds are expanded by RRT to 100{,}000 grasps, and it shares no grasp with the training dataset. In every real-robot experiment, a goal is reached if the position and orientation errors are both within the tolerance at any step of its window, with no hold or twist condition. The coarse tolerance is 2\,cm and 0.4\,rad, the precise one 1\,cm and 0.2\,rad, and the orientation error is the axis tilt for the lightbulb and wineglass. Filters run at $w = 0.5$ unless stated otherwise.

A drop is the object more than 25\,cm from the hand (goals lie within 20\,cm of it) for three consecutive tracked control steps (0.15\,s). This is tighter than the 0.3\,m safe set of simulation (Sec.~\ref{sec:problem}) because the table in the motion-capture cage can sometimes be less than 0.3\,m below the hand. After a drop, the hand opens to a fixed starting grasp, we place the object back in the hand, and the run continues with the next goal in the sequence. The object is stuck if two consecutive goals fail with the object having not moved (within 10\,mm and 10$^\circ$ over the last 5\,s). The hand is then reset the same way and the run continues with the next goal. If the hand loses its connection mid-goal, for example when the object hits the hand's USB cable, we reset the hand to the same starting grasp and redo that goal; the interrupted attempt is not counted.

With the additional wide-goal and lightbulb-twist tests (400 goals of 10\,s and 200 of 20\,s), the real-robot evaluation totals 5{,}400 goals and 15.6\,hours of goal windows. Fig.~\ref{fig:goal_outcomes} shows the outcome of every goal of the main experiment, run by run.

\begin{figure*}[t]
    \centering
    \includegraphics[width=\textwidth]{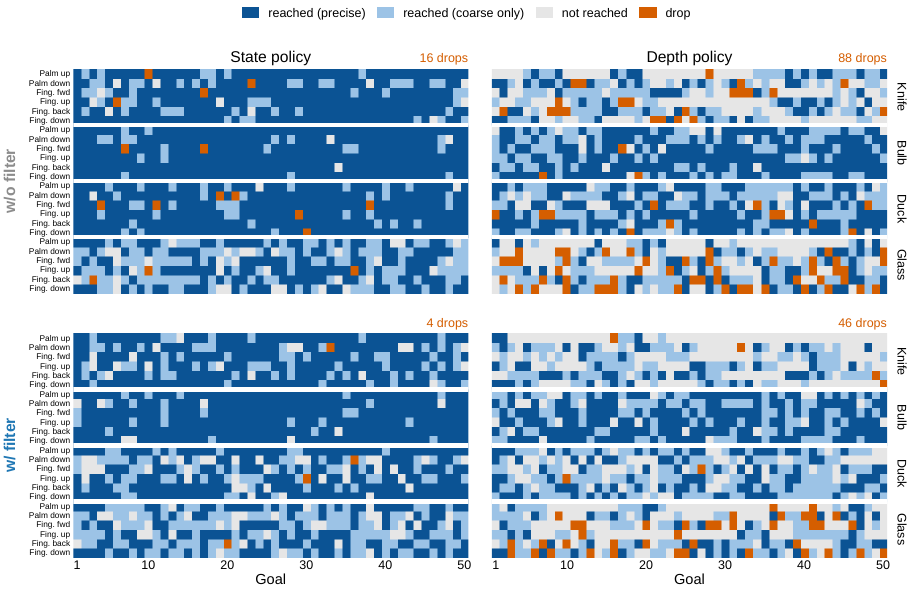}
    \caption{\textbf{Outcome of every goal in the main real-robot experiment (96 runs, 4{,}800 goals).} Each row of cells is one run of 50 goals. Dark blue: precise; light blue: coarse only; grey: not reached; orange: drop.}
    \label{fig:goal_outcomes}
\end{figure*}


\textbf{Wide goals.} The wide evaluation set (Table~\ref{tab:wide_goals}) uses the same pipeline (20{,}000 LightningGrasp seeds expanded by RRT to 100{,}000 grasps, generated independently), but with the canonical box enlarged 1.5$\times$ about its center, so the object can be placed farther from the palm. As before, the grasps pass the generator's force-balance and collision tests but are not physically validated in simulation or on the robot, so some goals may simply be unreachable. Of the goals we ran, 34--52\,\% of the wide goals lie inside the original canonical box, against 78--86\,\% of the main evaluation goals (Fig.~\ref{fig:goal_diversity}). 

\begin{figure}[t]
    \centering
    \includegraphics[width=\columnwidth]{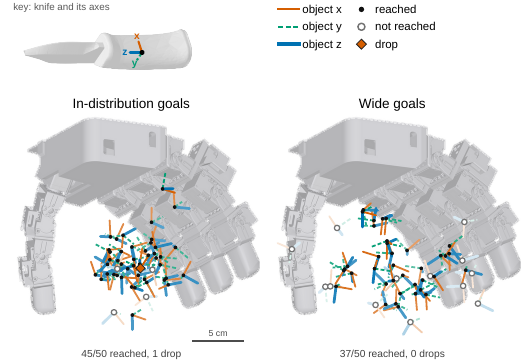}
    \caption{\textbf{Goal diversity, knife, palm down.} Goal poses of one run from the main (left) and wide (right) evaluation sets, marked reached (dot), not reached (open circle) or dropped (diamond).}
    \label{fig:goal_diversity}
\end{figure}

\begin{table}[t]
\centering
\scriptsize
\setlength{\tabcolsep}{3pt}
\caption{Wide goals from a 1.5$\times$ larger grasp box: state policy, palm down, 50 goals per setting, without ($\times$) and with ($\checkmark$) the filter. Avg.\ action change: RMS change of the action by the filter (\% of the action limit).}
\label{tab:wide_goals}
\begin{tabular}{@{}ll ccc c@{}}
\toprule
 & & \multicolumn{2}{c}{Goals reached (/50)\,$\uparrow$} & \makecell{Drops\\(/50)\,$\downarrow$} & \makecell{Avg.\\action\\change\\(\%)} \\
\cmidrule(lr){3-4}
Object & \makecell{Safety\\filter} & precise & coarse & &  \\
\midrule
\multirow{2}{*}{Knife} & $\times$ & 24 & 37 & 0 & -- \\
 & $\checkmark$ & 21 & 39 & 0 & 12.1 \\
\addlinespace[2pt]
\multirow{2}{*}{Lightbulb} & $\times$ & 30 & 42 & 0 & -- \\
 & $\checkmark$ & 23 & 38 & 0 & 11.9 \\
\addlinespace[2pt]
\multirow{2}{*}{Rubber duck} & $\times$ & 25 & 42 & 0 & -- \\
 & $\checkmark$ & 12 & 28 & 0 & 21.7 \\
\addlinespace[2pt]
\multirow{2}{*}{Wineglass} & $\times$ & 8 & 28 & 2 & -- \\
 & $\checkmark$ & 6 & 19 & 2 & 16.1 \\
\bottomrule
\end{tabular}
\end{table}

\textbf{Lightbulb spinning.} The depth policy and filter of the main experiment run palm up and palm down (Table~\ref{tab:lightbulb_twist}). Each run has 50 goals of 20\,s from the main evaluation set, each with a twist command of $+1$ or $-1$\,rad/s drawn at random. We count how many turns the bulb makes about its axis in the commanded direction while it is within the coarse tolerance.

\textbf{Arm motion.} This test (Table~\ref{tab:motion_r2}) uses the depth policies, since the large arm motions cannot be run inside the motion-capture cage. The lightbulb is given a goal and a twist command ($-1.5$\,rad/s, or $-1$\,rad/s for the high goal), and the rubber duck's goal rotates continuously about its $z$-axis at 0.5\,rad/s. The high goal is near the top of each object's goal range. A trial counts as held if the object is still in hand at the end. The filter's largest gains are at the high goals (from 7/10 to 10/10 trials held for the lightbulb, and from 7/10 to 9/10 for the duck). At the high lightbulb goal (15.6\,cm), the fingers barely moved in 3 of the 10 trials with the filter. We suspect that in these hand configurations, the filter corrects the policy strongly enough to stall it.

\textbf{Downstream tasks.} Table~\ref{tab:tasks} gives the results. Each task is a scripted sequence of arm motions and in-hand goals for the depth policy, run 10 times (5 for the lightbulb) without and with the filter at $w = 1.5$. In the rubber duck task, the hand grasps the duck lying on its side, lifts it, reorients it base-down (10\,s goal), then lowers it and opens the fingers. In the lightbulb task, it brings the bulb lying in the palm upright (8\,s goal), carries it under the socket, and screws it in with a $+1$\,rad/s twist goal until it lights. In the wineglass task, it grasps and lifts the glass, flips it 180$^\circ$, and flips it back. Without motion capture, we score each stage by eye; for example, the bulb counts as screwed in when it lights.

\subsection{Additional Results}
\label{app:results}

\textbf{Simulation results.} Table~\ref{tab:sim_results} gives the metrics behind Fig.~\ref{fig:sim_w_dial}. Without the filter, the depth policy reaches only 1--5 percentage points fewer goals than the state policy at the coarse tolerance, but 9--31 points fewer at the precise tolerance, likely because the object pose inferred from depth is less precise than the ground-truth pose the state policy observes.

\textbf{Real robot vs.\ simulation.} The state policies' coarse success is 96--99\,\% on the real robot against 92--95\,\% in simulation (Table~\ref{tab:sim_results}) for the knife, lightbulb and duck, and 89\,\% in both for the wineglass, and they drop 0.7--2.3\,\% of goals against 3.7--9.1\,\% in simulation. The two protocols differ, however: the simulation benchmark uses random wrist orientations, goals from the training distribution, and the heavy domain randomization of the simulation environment, while the real runs use six different wrist orientations and unseen goals. Fig.~\ref{fig:success_drop_trade_off} shows the trade-off per object, and Fig.~\ref{fig:more_filter_saves} shows more drops that the filter prevented.

\begin{table*}[t]
\centering
\scriptsize
\caption{Simulated results behind Fig.~\ref{fig:sim_w_dial} for every correction weight $w$ (1{,}024 environments $\times$ 50 goals). Coarse and precise: goals reached (\%) within 2\,cm, 0.4\,rad and 1\,cm, 0.2\,rad; drops: \% of goals; $T_\text{drop}$: mean time to the first drop (s; 500 if none). $^{\dagger}$Deployed on the real robot.}
\label{tab:sim_results}
\setlength{\tabcolsep}{2.6pt}
\begin{tabular*}{\textwidth}{@{\extracolsep{\fill}}l rrrr rrrr rrrr rrrr@{}}
\toprule
 & \multicolumn{4}{c}{Knife} & \multicolumn{4}{c}{Lightbulb} & \multicolumn{4}{c}{Rubber duck} & \multicolumn{4}{c}{Wineglass} \\
\cmidrule(lr){2-5}\cmidrule(lr){6-9}\cmidrule(lr){10-13}\cmidrule(lr){14-17}
Filter $w$ & Coarse & Precise & Drops & $T_\text{drop}$ & Coarse & Precise & Drops & $T_\text{drop}$ & Coarse & Precise & Drops & $T_\text{drop}$ & Coarse & Precise & Drops & $T_\text{drop}$ \\
\midrule
\multicolumn{17}{@{}l}{\textit{State policy (A3) and state filter (B1)}} \\
None & 94.8 & 91.2 & 3.7 & 255 & 93.8 & 92.1 & 6.5 & 266 & 92.1 & 88.4 & 9.1 & 154 & 88.6 & 67.7 & 6.1 & 287 \\
\addlinespace[1pt]
2 & 94.8 & 90.1 & 3.0 & 296 & 96.9 & 94.8 & 2.9 & 317 & 92.8 & 88.0 & 7.2 & 205 & 91.5 & 67.8 & 3.0 & 312 \\
1.5 & 94.5 & 89.4 & 2.9 & 304 & 97.0 & 95.0 & 2.9 & 315 & 92.6 & 87.0 & 6.6 & 208 & 91.3 & 67.2 & 2.6 & 323 \\
1 & 94.4 & 88.6 & 2.6 & 304 & 97.1 & 95.0 & 2.5 & 333 & 92.6 & 85.4 & 5.9 & 232 & 91.0 & 65.6 & 2.6 & 317 \\
0.5$^{\dagger}$ & 94.4 & 87.6 & 2.5 & 321 & 97.2 & 94.9 & 2.4 & 343 & 91.3 & 80.1 & 4.5 & 266 & 91.3 & 64.2 & 2.5 & 325 \\
0 & 94.3 & 85.7 & 2.1 & 330 & 97.3 & 94.6 & 2.2 & 349 & 70.4 & 53.2 & 2.5 & 351 & 90.1 & 60.9 & 2.4 & 328 \\
\midrule
\multicolumn{17}{@{}l}{\textit{Depth policy (A4) and depth filter (B2)}} \\
None & 89.7 & 61.8 & 4.4 & 221 & 93.1 & 82.8 & 6.4 & 197 & 87.0 & 57.4 & 10.2 & 148 & 84.0 & 50.3 & 4.4 & 238 \\
\addlinespace[1pt]
2 & 89.6 & 58.2 & 2.8 & 286 & 94.0 & 80.6 & 4.0 & 258 & 86.9 & 52.0 & 7.0 & 182 & 84.1 & 48.5 & 3.7 & 258 \\
1.5 & 89.2 & 57.7 & 2.8 & 286 & 93.8 & 79.9 & 3.7 & 257 & 87.1 & 50.5 & 5.9 & 201 & 84.4 & 48.1 & 3.5 & 268 \\
1 & 89.3 & 57.3 & 2.7 & 292 & 93.5 & 79.0 & 3.8 & 264 & 86.0 & 47.3 & 4.8 & 227 & 84.0 & 47.7 & 3.6 & 262 \\
0.5$^{\dagger}$ & 89.0 & 56.0 & 2.8 & 290 & 93.6 & 77.9 & 3.2 & 285 & 81.7 & 39.1 & 3.4 & 275 & 83.3 & 46.3 & 3.5 & 273 \\
0 & 88.6 & 55.2 & 2.8 & 281 & 93.2 & 76.8 & 3.2 & 292 & 51.8 & 18.0 & 1.3 & 372 & 83.0 & 44.9 & 3.1 & 277 \\
\bottomrule
\end{tabular*}
\end{table*}

\begin{figure}[t]
    \centering
    \includegraphics[width=\columnwidth]{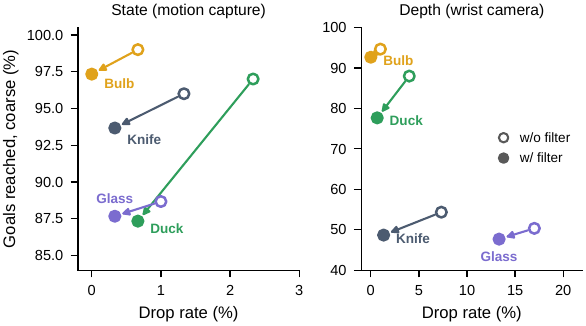}
    \caption{\textbf{Goal success versus drop rate on the real robot}, per object, from without (open) to with (filled) the safety filter, for the state (left) and depth (right) policies.}
    \label{fig:success_drop_trade_off}
\end{figure}

\begin{figure*}[t]
    \centering
    \includegraphics[width=\textwidth]{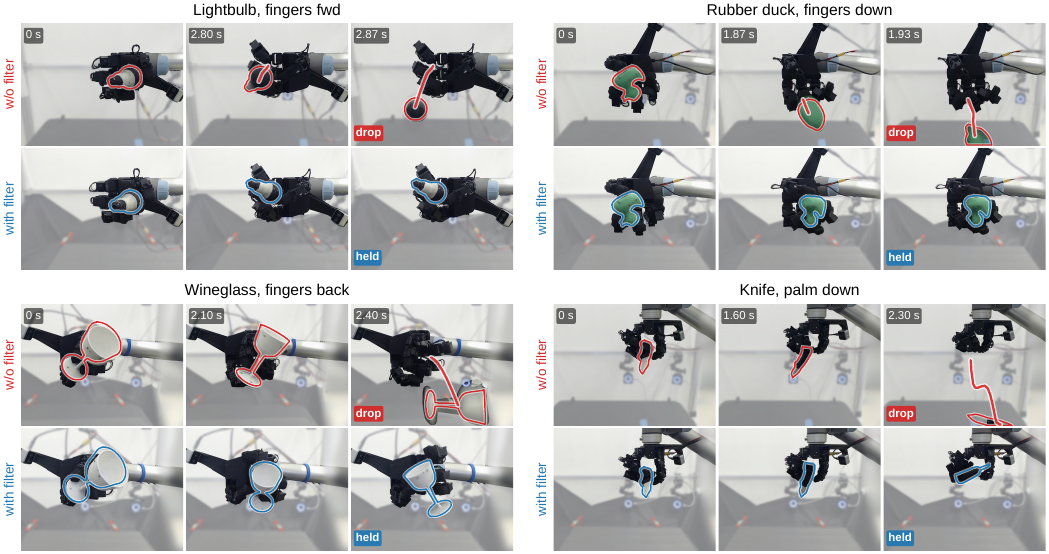}
    \caption{\textbf{More drops prevented by ResSafe.} The same goal without (top) and with (bottom) the filter; the red line traces the fall path.}
    \label{fig:more_filter_saves}
\end{figure*}

\subsubsection{Where Does the Depth Gap Come From?}
\label{app:depth_gap}
Since the state policies transfer well, the simulated dynamics seem adequate, and the depth observation is the only change. To test whether the gap is perceptual, we replay recorded real wrist stereo pairs through the stereo network and the depth policy, read out its auxiliary object-pose head, which predicts the object pose from the same features the policy acts on, and compare it with motion capture (Fig.~\ref{fig:aux_pose}, pooled over the state- and depth-policy runs of the main experiment). For the rubber duck and the lightbulb, the head is as accurate on the real robot as in simulation (median errors within about 1\,mm and 1$^\circ$ of simulation). For the knife and the wineglass, the real errors have long tails that simulation does not show: the orientation error exceeds 45$^\circ$ in 22 and 21\,\% of the real steps, against 5 and 0.1\,\% in simulation, and the wineglass's median position error grows from 10 to 16.5\,mm. This suggests a need for more accurate system identification of the depth sensing, or heavier visual augmentation and domain randomization in training.

\begin{figure}[t]
    \centering
    \includegraphics[width=\columnwidth]{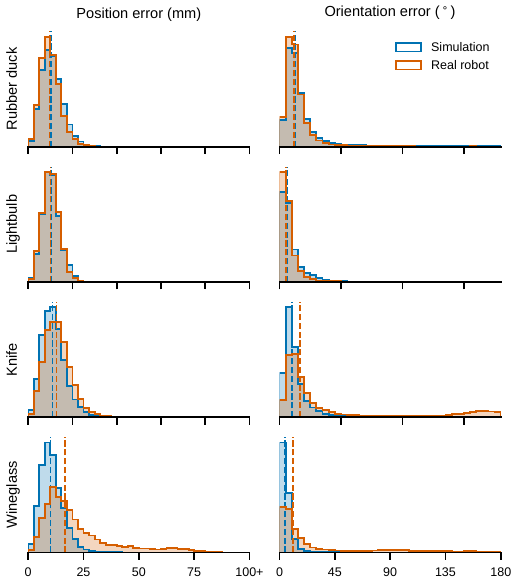}
    \caption{\textbf{Object pose perceived by the depth policy.} Error of its auxiliary pose head in simulation (blue) and on the real robot (orange); dashed lines are medians.}
    \label{fig:aux_pose}
\end{figure}


\begin{table}[t]
\centering
\scriptsize
\setlength{\tabcolsep}{3.5pt}
\caption{Lightbulb spinning: depth policy, 50 goals of 20\,s per row. Turns about the bulb axis in the commanded direction, counted within the coarse tolerance; $\geq\!\tfrac12$ / $\geq$1: goals with at least half / one full turn.}
\label{tab:lightbulb_twist}
\begin{tabular}{@{}ll cc ccc@{}}
\toprule
 & & & & \multicolumn{3}{c}{Turns\,$\uparrow$} \\
\cmidrule(l){5-7}
Wrist & \makecell{Safety\\filter} & Reached\,$\uparrow$ & Drops\,$\downarrow$ & $\geq\!\tfrac12$ & $\geq$1 & mean \\
\midrule
\multirow{2}{*}{Palm up} & $\times$ & \textbf{46} & 0 & \textbf{34} & \textbf{28} & \textbf{1.24} \\
 & $\checkmark$ & 43 & 0 & 25 & 20 & 0.74 \\
\addlinespace[2pt]
\multirow{2}{*}{Palm down} & $\times$ & \textbf{44} & 0 & \textbf{19} & \textbf{12} & \textbf{0.63} \\
 & $\checkmark$ & 41 & 0 & 14 & 7 & 0.36 \\
\bottomrule
\end{tabular}
\end{table}

\begin{table}[t]
\centering
\scriptsize
\setlength{\tabcolsep}{3.5pt}
\caption{Trials (of 10) with the object still in hand after 30\,s of arm motion. Depth policy, filter at $w{=}0.5$; goal height above the palm.}
\label{tab:motion_r2}
\begin{tabular}{@{}l cc cc@{}}
\toprule
 & & & \multicolumn{2}{c}{\makecell{Held (/10)\,$\uparrow$}} \\
\cmidrule(l){4-5}
 & \makecell{Goal height\\(cm)} & \makecell{Spin\\(rad/s)} & \makecell{No safety\\filter ($\times$)} & \makecell{Safety\\filter ($\checkmark$)} \\
\midrule
Lightbulb & 11.5 & $-1.5$ & 10/10 & 10/10 \\
Lightbulb & 15.6 & $-1.0$ & 7/10 & \textbf{10/10} \\
Rubber duck & 13.0 & 0.5 & 8/10 & \textbf{9/10} \\
Rubber duck & 16.5 & 0.5 & 7/10 & \textbf{9/10} \\
\midrule
All & & & 32/40 & \textbf{38/40} \\
\bottomrule
\end{tabular}
\end{table}

\begin{table}[t]
\centering
\scriptsize
\setlength{\tabcolsep}{3pt}
\caption{Downstream tasks: cumulative stage success (a trial passes a stage only if it passed all earlier ones). Depth policy without ($\times$) and with ($\checkmark$, $w{=}1.5$) the filter.}
\label{tab:tasks}
\begin{tabular}{@{}c ccc@{}}
\toprule
\makecell{Safety\\filter} & \multicolumn{3}{c}{Trials that passed each stage (cumulative)\,$\uparrow$} \\
\midrule
\multicolumn{4}{@{}l}{\textbf{Rubber duck}: grasp, reorient \& place} \\
 & Grasp\,+\,lift & \makecell{Reorient\\(base down)} & \makecell{Place\\upright} \\
\cmidrule(l){2-4}
$\times$ & \textbf{8/10} & \textbf{8/10} & \textbf{7/10} \\
$\checkmark$ & 7/10 & 6/10 & 3/10 \\
\midrule
\multicolumn{4}{@{}l}{\textbf{Lightbulb}: screw in} \\
 & \makecell{Upright\\from palm} & \makecell{Into\\socket} & \makecell{Screwed in\\(lights up)} \\
\cmidrule(l){2-4}
$\times$ & \textbf{5/5} & \textbf{4/5} & \textbf{4/5} \\
$\checkmark$ & \textbf{5/5} & 3/5 & 3/5 \\
\midrule
\multicolumn{4}{@{}l}{\textbf{Wineglass}: pick up \& flip} \\
 & Pick up & Flip 180$^\circ$ & Flip back \\
\cmidrule(l){2-4}
$\times$ & \textbf{10/10} & \textbf{8/10} & \textbf{7/10} \\
$\checkmark$ & \textbf{10/10} & \textbf{8/10} & 6/10 \\
\bottomrule
\end{tabular}
\end{table}

\subsubsection{Ablations}
\label{app:abl}
Figs.~\ref{fig:curriculum_ablation}--\ref{fig:ressafe_training} give the full results of the ablations of Sec.~\ref{sec:ablations}.

\textbf{Curricula.} Fig.~\ref{fig:curriculum_ablation} repeats the curriculum ablation on all four objects with one seed, comparing goals per episode at the end of A1 training. On the lightbulb and rubber duck, the friction curriculum gives 15--64\,\% more goals per episode than no curriculum. Goal and friction curricula together are the best combination on three of the four objects. The exception is the wineglass, where training without any curriculum ends highest at this budget, with 2.74 goals per episode against 2.20 with both curricula, so the benefit is not universal. The goal curriculum alone does not help, since it delays reaching full difficulty.

\textbf{Privileged actor observations.} Fig.~\ref{fig:training_observation_ablation} shows the A1 training curves on the knife with the privileged actor and with an actor that sees only the student's noisy, delayed observations.

\textbf{ResSafe design.} Fig.~\ref{fig:ressafe_ablation} shows the trade-off of the filter variants over $w$ in simulation. At $w = 0.5$, averaged over three seeds, the default L2 penalty reaches 80.4\,\% of goals at the coarse tolerance and drops 3.8\,\%. The L1 penalty drops fewer, 1.3\,\%, but reaches only 21.7\,\%. Without the nominal-action noise, the filter reaches 78.6\,\% and drops 2.5\,\%, close to the default. Fig.~\ref{fig:ressafe_training} shows B1 training: all three variants reach a similar rate of unsafe episodes, but the L1 penalty lets the filter suppress much more of the task, so it completes far fewer goals during training.

\begin{figure}[t]
    \centering
    \includegraphics[width=\columnwidth]{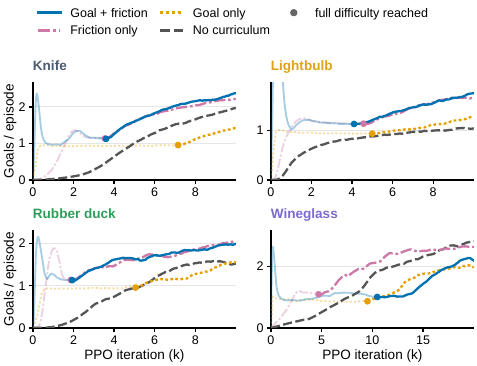}
    \caption{\textbf{Curriculum ablation for A1} on all four objects (one seed). Curves are faded while a curriculum is still ramping; dots mark full difficulty.}
    \label{fig:curriculum_ablation}
\end{figure}

\begin{figure}[t]
    \centering
    \includegraphics[width=\columnwidth]{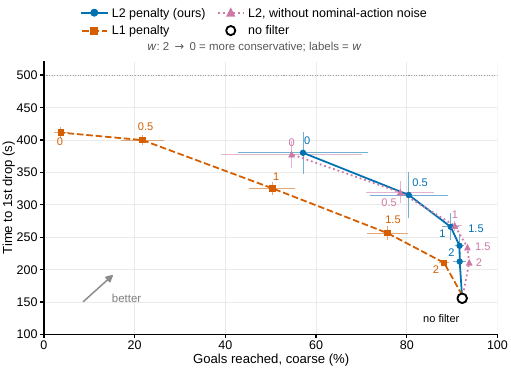}
    \caption{\textbf{ResSafe penalty ablation} (rubber duck, 3 seeds): L2 penalty (ours), L1 penalty, and L2 without nominal-action noise. Labels give $w$; error bars: s.e.m.\ over seeds.}
    \label{fig:ressafe_ablation}
\end{figure}

\begin{figure}[t]
    \centering
    \includegraphics[width=\columnwidth]{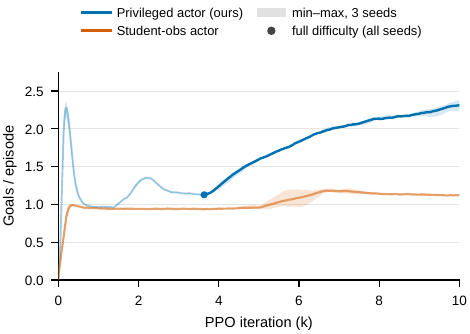}
    \caption{\textbf{A1 needs privileged actor observations.} Knife goals per episode with the privileged actor (ours) and with the student's observations (3 seeds).}
    \label{fig:training_observation_ablation}
\end{figure}

\begin{figure}[t]
    \centering
    \includegraphics[width=\columnwidth]{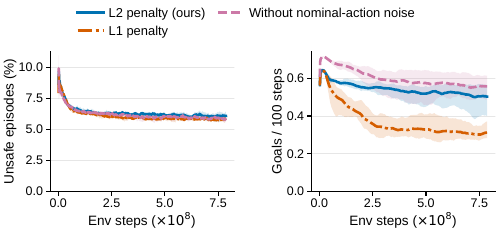}
    \caption{\textbf{ResSafe B1 training} (rubber duck, 3 seeds): unsafe-episode rate (left) and goals per 100 steps (right); mean with min--max band over seeds.}
    \label{fig:ressafe_training}
\end{figure}

\fi 

\end{document}